\documentclass[preprint,12pt]{elsarticle}

\usepackage[ruled]{algorithm2e} 
    
\usepackage[utf8]{inputenc}
\usepackage{amsmath,amsfonts,amssymb,amsthm}
	\numberwithin{equation}{section} 
\usepackage[table]{xcolor}
\usepackage{natbib}
\usepackage{enumitem}[shortlabels]
    \setitemize{noitemsep,topsep=0pt,parsep=0pt,partopsep=0pt}  
    \setenumerate{noitemsep,topsep=0pt,parsep=0pt,partopsep=0pt}  
\usepackage{wrapfig}
\usepackage{float}
\usepackage{thm-restate}
\usepackage[T1]{fontenc}    
	\usepackage{lmodern} 
\usepackage{hyperref}       
    \usepackage{url}            
\usepackage{booktabs}       
\usepackage{nicefrac}       
\usepackage{microtype}      

\DeclareMathAlphabet{\mathpzc}{OT1}{pzc}{m}{it}

\newtheorem{theorem}{\protect\theoremname}[section]
\newtheorem{definition}[theorem]{\protect\definitionname}
\newtheorem{lemma}[theorem]{\protect\lemmaname}
\newtheorem{proposition}[theorem]{\protect\propositionname}
\newtheorem{example}[theorem]{\protect\examplename}
\newtheorem{corollary}[theorem]{\protect\corollaryname}
\newtheorem{remark}[theorem]{\protect\remarkname}

\providecommand{\corollaryname}{Corollary}
\providecommand{\claimname}{Claim}
\providecommand{\definitionname}{Definition}
\providecommand{\lemmaname}{Lemma}
\providecommand{\notationname}{Notation}
\providecommand{\remarkname}{Remark}
\providecommand{\problemname}{Problem}
\providecommand{\propositionname}{Proposition}
\providecommand{\examplename}{Example}
\providecommand{\theoremname}{Theorem}

\newcommand{\N}{\mathbb{N}}
\newcommand{\R}{\mathbb{R}}
\newcommand{\mc}{\mathcal}
\newcommand{\mb}{\mathbb}

\newcommand{\defword}[1]{\textbf{\boldmath{#1}}}
\newcommand{\pub}{s_{\textnormal{pub}}}
\newcommand{\pubTree}{\mathcal S_{\textnormal{pub}}}
\newcommand{\pubSet}{I_{\textnormal{pub}}}
\newcommand{\pubSetTree}{\mc I_{\textnormal{pub}}}
\newcommand{\modelAbbrev}{FOSG}
\newcommand{\modelName}{factored-observation stochastic game}

\newcommand{\ModelNameCamel}{Factored-Observation Stochastic Game}
\newcommand{\initState}{w^{\textnormal{init}}}
\newcommand{\EF}[1]{\texttt{ExtForm}\left(#1\right)}
\newcommand{\ClassicalEFG}[1]{\texttt{ClassicalEFG}\left(#1\right)}

\newcounter{vkNoteCounter}

\newcommand{\edit}[1]{\textcolor{purple}{#1}}
\renewcommand{\edit}[1]{#1}              
\usepackage{graphicx}

\usepackage{xcolor}
\usepackage{forest} 
	\usetikzlibrary{shapes} 
	\usetikzlibrary{calc} 
\usepackage{xparse}	

\newlength{\nodesize}
\colorlet{chance_color}{black}
\colorlet{pl0_color}{chance_color}
\colorlet{chance_text}{white}
\colorlet{pl1_color}{magenta!50}
\colorlet{pl2_color}{cyan!50}
\colorlet{pl0_infoset_color}{pl0_color}
\colorlet{pl1_infoset_color}{magenta!75}
\colorlet{pl2_infoset_color}{cyan!75}
\forestset{
	basenode/.style = {draw,
		inner sep = 0,
		outer ysep = 0,
		minimum size = \nodesize,
		anchor = north
	},
	playernode/.style={basenode, 
		shape = regular polygon,
		regular polygon sides = 3,
	},
	pl1/.style={playernode, fill=pl1_color},
	pl2/.style={playernode, fill=pl2_color, shape border rotate=180},
	chance/.style = {basenode,
		fill=pl0_color, text=chance_text,
		circle,
		minimum size=0.75*\nodesize,
		l sep=0.4765\nodesize,
	},
	terminal/.style = {basenode,
		shape = regular polygon,
		regular polygon sides = 4,
		l sep=0.47\nodesize,
		minimum size = 1.07\nodesize
	}
}
\tikzset{
	partition/.style = {
		draw,
		rounded corners = 5,
		inner sep=0.1\nodesize,
	},
	infoset/.style = {
		partition,
		draw=pl1_infoset_color,
	},
	augmented/.style = {
		dashed
	},
	opponent/.style = {
		draw=pl2_infoset_color,
		inner sep=0.225\nodesize,
	},
	pl1_cl_infoset/.style = {infoset, yshift=-0.035\nodesize},
	pl2_cl_infoset/.style = {infoset,
		opponent,
		inner sep=0.1\nodesize,
		yshift=0.035\nodesize
	},
	public_state/.style = {
		partition,
		draw=pl0_infoset_color,
		inner sep=0.125\nodesize,
		rounded corners = 7,
	},
}
\tikzset{
	line_infoset/.style = {
		draw=pl1_infoset_color,
		dashed
	},
	opp_line_infoset/.style = {
		draw=pl2_infoset_color,
		dashed
	}
}
\newcommand{\corners}[1]{#1.corner 1)(#1.corner 2)(#1.corner 3}

\pgfdeclarelayer{bg}    
\pgfsetlayers{bg,main}  
    \usepackage{graphicx}
    \usepackage{caption}
    \usepackage{subcaption}
    \usepackage{tikz} 
    	\usetikzlibrary{shapes} 
    	\usetikzlibrary{calc} 
    	\usepackage{forest}
    \usepackage{xparse}	
    	\usetikzlibrary{shapes.multipart} 
    	\usetikzlibrary{decorations.pathmorphing} 
    	\usetikzlibrary{decorations.pathreplacing} 
    \colorlet{chance_color}{black}
    \colorlet{pl0_color}{chance_color}
    \colorlet{chance_text}{white}
    \colorlet{pl1_color}{magenta!50}
    \colorlet{pl2_color}{cyan!50!lime!60}
    \tikzset{
    	basenode/.style = {draw,
    		inner sep = 0,
    		minimum size = \nodesize
    	},
    	playernode/.style={basenode, 
    		shape = regular polygon,
    		regular polygon sides = 3
    	},
    	pl1/.style={playernode, fill=pl1_color},
    	pl2/.style={playernode, fill=pl2_color, shape border rotate=180},
    	chance/.style = {basenode,
    		fill=pl0_color, text=chance_text,
    		circle,
    		minimum size=0.7*\nodesize,
    	},
    	terminal/.style = {basenode,
    		draw=none,
    		outer sep=0,
    		minimum size = 0.6\nodesize
    	}
    }
    \NewDocumentCommand{\IS}{ m m O{0.6} O{0.6} O{0.6} O{5} O{dashed} O{} O{1} }{
    	\draw [#7, rounded corners=#6, draw=pl#9_color!75!black]
    		($(#1)+(-#5\nodesize,#3\nodesize)$) rectangle
    		($(#2)+(#5\nodesize,-#4\nodesize)$);
    	\node at ($(#1)!.5!(#2)$) {#8};
    }
    \NewDocumentCommand{\oneNodeIS}{ m O{0.6} O{0.6} O{0.6} O{5} O{dashed} O{} O{1} }{
    	\IS{#1}{#1}[#2][#3][#4][#5][#6][#7][#8]
    }
    \NewDocumentCommand{\actIS}{ m m O{0.6} O{0.6} O{0.6} O{5} O{} O{1} }{
    	\IS{#1}{#2}[#3][#4][#5][#6][solid][#7][#8]
    }
    \NewDocumentCommand{\oppActOneNodeIS}{ m O{0.6} O{0.6} O{0.6} O{5} O{solid} O{} }{
    	\IS{#1}{#1}[#2][#3][#4][#5][#6][#7][2]
    }
    \NewDocumentCommand{\oppIS}{ m m O{0.6} O{0.6} O{0.6} O{5} O{dashed} O{} }{
    	\IS{#1}{#2}[#3][#4][#5][#6][#7][#8][2]
    }
    \NewDocumentCommand{\publicState}{ m m O{0.6} O{0.6} O{0.6} O{0} O{}}{
    	\IS{#1}{#2}[#3][#4][#5][#6][solid][#7][0]
    }
    \tikzset{
    	level 1/.style = {level distance=2\nodesize, sibling distance=3\nodesize},
    	level 2/.style = {level distance=2\nodesize, sibling distance=1.5\nodesize}
    }
    \def\infosetDeprecated[#1,#2,#3](#4,#5)(#6,#7,#8)(#9){	
    	\draw [#1, rounded corners=#2, draw=pl#3_color!75!black]
    		($(#4)+(-#6\nodesize,#7\nodesize)$) rectangle
    		($(#5)+(#6\nodesize,-#8\nodesize)$);
    	\node at ($(#4)!.5!(#5)$) {#9};
    }
    \def\actInfosetDeprecated(#1,#2){
    	\infosetDeprecated[solid,5,1](#1,#2)(0.6,0.6,0.5)()
    }
    \def\augInfosetDeprecated(#1,#2){
    	\infosetDeprecated[dashed,5,1](#1,#2)(0.6,0.5,0.6)()
    }

    \def\sneakingGamePartition{
    \setlength{\nodesize}{1.85em}
    \tikzset{
    	level 1/.style = {level distance=2\nodesize, sibling distance=3\nodesize},
    	level 2/.style = {level distance = 1.5\nodesize, sibling distance=1.5\nodesize},
    	level 3/.style = {level distance = 1.5\nodesize},
    	level 4/.style = {level distance = 1.75\nodesize}
    }
    \node(root)
    [chance,label=right:{}]{}
    	child{node(H1)[pl1]{}
    		child{node{}
    		edge from parent[draw=none]{}
    			child{node(H2)[pl1]{}
    				edge from parent[draw=none]{}
    				child{node[terminal]{$\frac{1}{2}$}
    					edge from parent node[left]{}
    				}							
    				child{node[terminal]{$1$}
    					edge from parent node[right]{}
    				}
    			}
    		}
    		edge from parent node[left]{$0.5$}
    		edge from parent node[left, yshift=0.5em, align=left]{}
    	}
    	child{node(L1)[pl1]{}
    		child{node(N)[pl2,label={[xshift=2.5em, yshift=-.0em]}]{$h_0$}
    			child[grow=-45]{node[terminal]{0}
    				edge from parent node[above right]{}
    			}
    			child[grow=down]{node(L2)[pl1]{}
    				child{node[terminal]{$\frac{1}{2}$}
    					edge from parent node[left]{}
    				}							
    				child{node[terminal]{$-1$}
    					edge from parent node[right]{}
    				}
    				edge from parent node[above left]{}
    			}
    			edge from parent node[left]{}
    		}
    		edge from parent node[right]{$0.5$}
    		edge from parent node[xshift=1.5em, yshift=0.5em, align=left]{}	}
    ;
    \draw (H1)--(H2);
    
    \draw[dashed, bend left, color=pl1_color!50!black] (H1) to (L1);
    \draw[dashed, bend left, color=pl1_color!50!black] (H2) to (L2);
    }
    
    \def\noFinestPartitionExample{
    \tikzset{
    	level 1/.style = {level distance=0\nodesize, sibling distance=2.1\nodesize},
    	level 2/.style = {level distance=1.25\nodesize},
    	level 3/.style = {level distance=1.5\nodesize},
    	level 4/.style = {level distance = 1.75\nodesize},
    	level 5/.style = {level distance = 0.9\nodesize}
    }
    \node(anchor){}
    child{node(A)[pl1]{}
    	edge from parent[draw=none]
    	child{node[pl2]{}
    		child{node[pl2]{}
    			child{node[missing]{}
    				child{node[pl1]{}
    					edge from parent[draw=none]
    				}			
    				edge from parent[draw=none]
    			}
    		}
    	}
    }
    child{node(B)[pl1, xshift=-0.8\nodesize]{}
    	edge from parent[draw=none]
    	child{node[pl2]{}
    		child{node[pl2]{}
    			child[grow=down]{node[missing]{}
    				child{node[pl1]{}
    					edge from parent[draw=none]
    				}			
    				edge from parent[draw=none]
    			}
    			child[grow=south east]{node[pl2,xshift=0.125\nodesize]{}
    				child[grow=down, level distance = 1.41\nodesize] {node[pl1]{}}
    			}
    		}
    	}
    }
    child{node(C)[pl1, xshift=0.8\nodesize]{}
    	edge from parent[draw=none]
    	child{node[pl2]{}
    		child{node[pl2]{}
    			child[grow=down]{node[missing]{}
    				child{node[pl1]{}
    					edge from parent[draw=none]
    				}			
    				edge from parent[draw=none]
    			}
    			child[grow=south west]{node[pl2, xshift=-0.125\nodesize]{}
    				child[grow=down, level distance = 1.41\nodesize] {node[pl1]{}}
    			}
    		}			
    	}
    }
    child{node(D)[pl1]{}
    	edge from parent[draw=none]
    	child{node[pl2]{}
    		child{node[pl2]{}
    			child{node[missing]{}
    				child{node[pl1]{}
    					edge from parent[draw=none]
    				}			
    				edge from parent[draw=none]
    			}
    		}
    	}
    };
    
    \foreach \L in {A,B,...,D}{
    	\draw (\L-1-1) -- (\L-1-1-1-1);
    	\draw[dotted] (\L) -- ($(\L)+(0,1.2\nodesize)$);
    	\draw[dotted] (\L-1-1-1-1) -- ($(\L-1-1-1-1)-(0,0.9\nodesize)$);
    }
    \draw[dotted] (B-1-1-2-1) -- ($(B-1-1-2-1)-(0,0.9\nodesize)$);
    \draw[dotted] (C-1-1-2-1) -- ($(C-1-1-2-1)-(0,0.9\nodesize)$);
    
    \actInfosetDeprecated(A,D)
    \actInfosetDeprecated(A-1-1-1-1,B-1-1-1-1)
    \actInfosetDeprecated(C-1-1-1-1,D-1-1-1-1)
    \actInfosetDeprecated(B-1-1-2-1,C-1-1-2-1)
    }
    
    \def\noFinestPartitionExampleA{
    \infosetDeprecated[dashed,5,1](A-1,D-1)(0.65,0.5,0.6)()
    \infosetDeprecated[dashed,5,1](A-1-1,D-1-1)(0.65,0.5,0.6)()
    \infosetDeprecated[dashed,5,1](B-1-1-2,C-1-1-2)(0.65,0.5,0.6)()
    }
    
    \def\noFinestPartitionExampleB{
    \draw [dashed,rounded corners, color=pl1_color!75!black]
    	($(A-1)+(-0.6\nodesize,0.5\nodesize)$)--($(D-1)+(0.6\nodesize,0.5\nodesize)$)--($(D-1-1)+(0.6\nodesize,-0.8\nodesize)$)--($(C-1-1)+(-0.55\nodesize,-0.8\nodesize)$)--($(C-1)+(-0.55\nodesize,-0.6\nodesize)$)--($(A-1)+(-0.6\nodesize,-0.6\nodesize)$)--cycle;
    \infosetDeprecated[dashed,5,1](A-1-1,C-1-1-2)(0.65,0.5,0.55)()
    }
    
    \def\perfRecallGame[#1]{
    \tikzset{
    	level 1/.style = {level distance=#1\nodesize, sibling distance=5\nodesize},
    	level 2/.style = {level distance=1.25\nodesize,sibling distance=2.5\nodesize},
    	level 3/.style = {level distance=1.35\nodesize, sibling distance=1.25\nodesize},
    	level 4/.style = {level distance=0.75\nodesize},
    }
    \node(R)[pl2]{}
    	child foreach[count=\i] \actorI/\actorII in {chance/pl1, pl1/pl2}{node[\actorI]{}
    		child foreach \j in {1,2}{node[\actorII]{}
    			child foreach \k in {1,2}{
    				node[terminal]{$u_{\i\j\k}$}
    			}
    		}
    	}
    ;
    }

    \def\pokerHistory{
    \tikzset{
    	level 1/.style = {level distance=1.25\nodesize, sibling distance=1.5\nodesize},
    	level 2/.style = {level distance=1.05\nodesize, sibling distance=1.5\nodesize},
    	level 3/.style = {level distance=1.25\nodesize, sibling distance=1.5\nodesize},		
    }
    
    \node(R0)[chance]{}
    \foreach \round in {1,2,3}{
    	child{node[pl1]{}
    		child{node[pl2]{}
    			child{node(R\round)[chance]{}}
    		}
    	}
    }
    	child{node[pl1]{}
    		child{node[pl2]{}
    			child{node(R4)[terminal]{end}}
    		}
    	}
    ;
    }

    \def\pokerAA{
    \foreach \i in {0,1,2,3}{
    	\draw[dashed] (R\i-1)--($(R\i-1)+(-.45\nodesize,-.6\nodesize)$);
    	\draw[dashed] (R\i-1)--($(R\i-1)+(.45\nodesize,-.6\nodesize)$);
    	\draw[dashed] (R\i-1-1)--($(R\i-1-1)+(-.45\nodesize,-.6\nodesize)$);
    	\draw[dashed] (R\i-1-1)--($(R\i-1-1)+(.45\nodesize,-.6\nodesize)$);
    	\draw[<-, dotted, bend left] ($(R\i.north east)+(0.05\nodesize,0.05\nodesize)$) to ($(R\i.north)+(1.0\nodesize,0)$);
    }
    \foreach \i in {1,2,3}{
        \draw[dashed] (R\i)--($(R\i)+(-.6\nodesize,-.4\nodesize)$);
    	\draw[dashed] (R\i)--($(R\i)+(-.45\nodesize,-.6\nodesize)$);
    	\draw[dashed] (R\i)--($(R\i)+(-.225\nodesize,-.7\nodesize)$);
    	\draw[dashed] (R\i)--($(R\i)+(.6\nodesize,-.4\nodesize)$);
    	\draw[dashed] (R\i)--($(R\i)+(.45\nodesize,-.6\nodesize)$);
    	\draw[dashed] (R\i)--($(R\i)+(.225\nodesize,-.7\nodesize)$);
    }
    \node[right=1.0\nodesize, align=left, rotate=-45] at (R0) {private\\cards dealt};
    \node[right=1.0\nodesize, align=left, rotate=-45] at (R0-1) {pl. 1 calls};
    \node[right=1.0\nodesize, align=left, rotate=-45] at (R0-1-1) {pl. 2 checks};
    \node[right=1.0\nodesize, align=left, rotate=-45] at (R1) {1st round\\ public cards};
    \node[right=1.0\nodesize, align=left, rotate=-45] at (R1-1) {pl. 1 checks};
    \node[right=1.0\nodesize, align=left, rotate=-45] at (R1-1-1) {pl. 2 checks};
    \node[right=1.0\nodesize, align=left, rotate=-45] at (R2) {2nd round\\ public cards};
    \node[right=1.0\nodesize, align=left, rotate=-45] at (R2-1) {pl. 1 checks};
    \node[right=1.0\nodesize, align=left, rotate=-45] at (R2-1-1) {pl. 2 checks};
    \node[right=1.0\nodesize, align=left, rotate=-45] at (R3) {last round\\ public cards};
    \node[right=1.0\nodesize, align=left, rotate=-45] at (R3-1) {pl. 1 checks};
    \node[right=1.0\nodesize, align=left, rotate=-45] at (R3-1-1) {pl. 2 ch.};
    \node at ($(R4)+(0,-0.25\nodesize)$) {};
    }
    
    \def\pokerB{
    \oneNodeIS{R0}[.5][0.5]
    \foreach \i/\iplusplus in {0/1, 1/2, 2/3, 3/4}{
    	\actIS{R\i-1}{R\i-1}[.6][0.5]
    	\IS{R\i-1-1}{R\iplusplus}[.45][0.45]
    }
    }
    
    \def\pokerC{
    \foreach \i in {0,1,2,3}{
    	\oppIS{R\i}{R\i-1}[.5][0.45]
    	\oppActOneNodeIS{R\i-1-1}[.5][0.6]
    }
    \oppActOneNodeIS{R4}[.5][0.45][0.6][5][dashed]
    }

    \def\pokerDD{
    \publicState{R0}{R0-1}[.5][0.45]
    \foreach \i/\iplusplus in {0/1, 1/2, 2/3}{
    	\publicState{R\i-1-1}{R\iplusplus-1}[.5][0.45]
    }
    \publicState{R3-1-1}{R4}[.5][0.45]
    \foreach \i/\description in {0/{pre-flop},1/$\ \ \ $flop,2/$\ \ \ $turn,3/$\ \ $river}{
    	\draw [decorate, decoration = {brace, amplitude=0.5\nodesize, raise=0.7\nodesize}]
    		($(R\i)+(0,0.25\nodesize)$) --
    		($(R\i-1-1)+(0,-0.45\nodesize)$);
    	\node[anchor=west, rotate=-90, align=center] at ($(R\i)!.5!(R\i-1-1)+(1.55\nodesize,0.8\nodesize)$) {\description};
    }
    }
    
    \def\unfairMP{
    \node(R)[chance]{}
    	child foreach \i in {1,2}{
    		node[pl\i]{}
    		child foreach \j in {1,2}{
    			node{}
    			child foreach \k in {1,2}{
    				node[terminal]{}
    			}
    		}
    	}
    ;
    \foreach \i/\player in {1/2, 2/1}{
    	\foreach \j in {1,2}{
    		\node[pl\player] at (R-\i-\j) {};
    	}
    }
    \foreach \coords/\payoff in {1-1-1/1, 1-1-2/-1, 1-2-1/-1, 1-2-2/1, 2-1-1/-1, 2-1-2/1, 2-2-1/1, 2-2-2/-1}{
    	\node at (R-\coords) {\payoff};
    }
    }
    \tikzset{
    	level 1/.style = {level distance=1.5\nodesize, sibling distance=5\nodesize},
    	level 2/.style = {level distance=2.25\nodesize, sibling distance=2\nodesize},
    	level 3/.style = {level distance=1.5\nodesize, sibling distance=1.\nodesize}
    }
    \def\unfairMPtext{
    \draw[rounded corners=15, solid, color=pl1_color!75!black, rotate around={-30:($(R-1)!.5!(R-2-1)$)}]
    	($(R-1)!.5!(R-2-1)+(-3\nodesize,0.6\nodesize)$) rectangle
    	($(R-1)!.5!(R-2-1)+(3\nodesize,-0.6\nodesize)$);
    \draw[rounded corners=15, solid, color=pl2_color!75!black, rotate around={30:($(R-2)!.5!(R-1-2)$)}]
    	($(R-2)!.5!(R-1-2)+(-3\nodesize,0.6\nodesize)$) rectangle
    	($(R-2)!.5!(R-1-2)+(3\nodesize,-0.6\nodesize)$);
    
    \node[xshift=-0.5\nodesize, rotate=+29.5] at ($(R)!.3!(R-1)$) {pick on pl.1};
    \node[xshift=+0.5\nodesize, rotate=-30.5] at ($(R)!.3!(R-2)$) {pick on pl.2};
    \node[xshift=-0.35\nodesize] at ($(R-1)!.3!(R-1-1)$) {H};
    \node[xshift=+0.35\nodesize] at ($(R-1)!.3!(R-1-2)$) {T};
    \node[xshift=-0.35\nodesize] at ($(R-2)!.3!(R-2-1)$) {T};
    \node[xshift=+0.35\nodesize] at ($(R-2)!.3!(R-2-2)$) {H};
    \foreach \i in {1,2}{
    	\foreach \j in {1,2}{
    		\node[xshift=-0.35\nodesize] at ($(R-\i-\j)!.5!(R-\i-\j-1)$) {H};
    		\node[xshift=+0.35\nodesize] at ($(R-\i-\j)!.5!(R-\i-\j-2)$) {T};
    	}
    }
    }

\def\quadraticPaddingGameA{
    \forestset{
	    dotz/.style = {
            edge label={node[midway, fill=white]{$\dots$}},
            edge={sloped}
    	},
    		dummy/.style = {basenode,
    		circle,
    		dotted,
    		minimum size=0.7*\nodesize,
    		calign=fixed edge angles
    	}
    }
    \centering
    \setlength{\nodesize}{2.5em}
    \begin{forest}
        [,pl2,name=h1
        [,phantom, tier=2,
            [,phantom, tier=3
                [,phantom, dotz, tier=N-1,
                    [, pl1, name=g1, s sep-=1em, tier=bottom [1] [0] ]
                    [,phantom]
                ]
                [,phantom]
            ]
            [,phantom]
            ]
        [, pl2, tier=2, name=h2
            [,phantom, tier=3,
                [,phantom, dotz, tier=N-1
                    [, pl1, s sep-=1em, tier=bottom, name=g2 [1] [0] ]
                    [,phantom]
                ]
                [,phantom]
            ]
            [, pl2, tier=3, name=h3
                [,phantom, dotz, tier=N-1
                    [, pl1, s sep-=1em, tier=bottom, name=g3 [1] [0] ]
                    [$\dots$, edge={draw=none}, tier=bottom, yshift=-1.25em]
                ]
                [, pl2, dotz, tier=N-1, name=h,
                    [, pl1, name=g, s sep-=1em, tier=bottom [1] [0] ]
                    [, pl1, name=R, s sep-=1em, tier=bottom [0] [1] ]
                ]
            ]
        ]
    ]
    \draw (h1) -- (g1);
    \draw (h2) -- (g2);
    \draw (h3) -- (g3);
    \node [pl1_cl_infoset, fit=(\corners{g1})(\corners{R})] {};
    \end{forest}
}    
\def\quadraticPaddingGameB{
    \forestset{
	dummy/.style = {basenode,
    		circle,
    		dotted,
    		minimum size=0.7*\nodesize,
    		calign=fixed edge angles
    	},
    dotz/.style = {
            edge label={node[midway]{$\dots$}},
            edge={sloped,draw=none}
    	}
    }
    \centering
    \setlength{\nodesize}{2.5em}
    \begin{forest}
    [$h_1$,pl2,
        [,chance, tier=2,
            [,chance, tier=3
                [,chance, dotz, tier=N-1,
                    [$g_1$, pl1, name=L, s sep-=1em, tier=bottom [1] [0] ]
                    [,phantom]
                ]
                [,phantom]
            ]
            [,phantom]
            ]
        [$h_2$, pl2, tier=2
            [,chance, tier=3,
                [,chance, dotz, tier=N-1
                    [$g_2$, pl1, s sep-=1em, tier=bottom [1] [0] ]
                    [,phantom]
                ]
                [,phantom]
            ]
            [$h_3$, pl2, tier=3,
                [,chance, dotz, tier=N-1
                    [$g_3$, pl1, s sep-=1em, tier=bottom [1] [0] ]
                    [$\dots$, edge={draw=none}, tier=bottom, yshift=-1.25em]
                ]
                [, pl2, dotz, tier=N-1, name=h,
                    [, pl1, name=g, s sep-=1em, tier=bottom [1] [0] ]
                    [, pl1, name=R, s sep-=1em, tier=bottom [0] [1] ]
                ]
            ]
        ]
    ]
    \node [pl1_cl_infoset, fit=(\corners{L})(\corners{R})] {};
    \node [xshift=0.3\nodesize, yshift=0.6\nodesize] at (h) {$h_{N-1}$};
    \node [xshift=-0.3\nodesize, yshift=0.8\nodesize] at (g) {$g_{N-1}$};
    \node [xshift=0.2\nodesize, yshift=0.8\nodesize] at (R) {$g_{N}$};
    \end{forest}
}

\usepackage{todonotes}

\journal{Artificial Intelligence}

\begin{document}

\begin{frontmatter}




\author{Vojtěch Kovařík\fnref{label1}}
\author{Martin Schmid\fnref{label2}}
\author{Neil Burch\fnref{label2}}
\author{\\Michael Bowling\fnref{label2}}
\author{Viliam Lisý\corref{cor1}\fnref{label1}}
\fntext[label1]{AI Center, FEE, Czech Technical University in Prague (\textit{Technická 2, Prague, 166 27, Czech Republic})}
\fntext[label2]{DeepMind, Alberta, Edmonton}
\cortext[cor1]{The correspondence should be addressed to lisyvili@fel.cvut.cz.}

\title{Rethinking Formal Models of Partially Observable Multiagent Decision Making}

\begin{abstract}

Multiagent decision-making in partially observable environments is usually modelled as either an extensive-form game (EFG) in game theory or a partially observable stochastic game (POSG) in multiagent reinforcement learning (MARL).
One issue with the current situation is that while most practical problems can be modelled in both formalisms, the relationship of the two models is unclear, which hinders the transfer of ideas between the two communities.
A second issue is that while EFGs have recently seen significant algorithmic progress, their classical formalization is unsuitable for efficient presentation of the underlying ideas, such as those around decomposition.

To solve the first issue, we introduce \modelName{}s (\modelAbbrev{}s), a minor modification of the POSG formalism which distinguishes between private and public observation and thereby greatly simplifies decomposition.
To remedy the second issue, we show that \modelAbbrev{}s and POSGs are naturally connected to EFGs: by ``unrolling'' a \modelAbbrev{} into its tree form, we obtain an EFG. Conversely, any perfect-recall timeable EFG corresponds to some underlying \modelAbbrev{} in this manner.
Moreover, this relationship justifies several minor modifications to the classical EFG formalization that recently appeared as an implicit response to the model's issues with decomposition.
Finally, we illustrate the transfer of ideas between EFGs and MARL by presenting three key EFG techniques -- counterfactual regret minimization, sequence form, and decomposition -- in the \modelAbbrev{} framework.

\end{abstract}




\end{frontmatter}


\section{Introduction}\label{sec:intro}

Sequential decision-making is one of the core topics of artificial intelligence research.
The ability of an artificial agent to perform actions, observe their consequences, and then perform further actions to achieve a goal is instrumental in domains from robotics and autonomous driving to medical decision diagnosis and automated personal assistants.
Recent progress has led to unprecedented results in many large-scale problems of this type. 
While conceptually simpler problems can be modelled with perfect information or by regarding the other agents as stationary parts of the environment, realistic models of real-world situations require rigorous treatment of imperfect information and multiple independent decision-makers operating in a shared environment.
The most popular game-theoretical model in these setting -- extensive form games (EFG) -- dates back to 1953 \citep{morgenstern1953theory}.
EFGs have served the community well, and many impressive results build on top of this particular framework \cite{DeepStack,Libratus,Pluribus}.

However, EFGs lose crucial information inherently present in many environments -- the notion of observations \edit{received by} the agents.
\edit{Observations are} essential not only for specifying \textit{what} \edit{information was received}, but also to know \textit{who} \edit{received} it and \textit{when}.
Since EFGs simply group states indistinguishable to the acting player, the notion of \edit{information being public or private} is forever lost, and so \edit{are the data about the timing}.
However, these concepts are essential for recent search algorithms \cite{DeepStack,brown2017safe,ReBeL}, where decomposition and reasoning about subgames crucially rely on the notion of public information.
While it is common to try to recover the necessary information from the EFG model \citep{CFR-D,accelerated_BR,MCCR}, we show that this is impossible to do in general.
Practical implementations thus bypass the model by using algorithms that are built with game-specific concepts (e.g., dealing cards in poker), rather than developing algorithms running purely on top of the information provided by the formal model \citep{DeepStack, Libratus}.

We argue that by a minor modification of another popular model -- partially-observable stochastic game (POSG) \cite{POSGs2004hansen} -- we can naturally describe domains in a way that preserves this necessary information.
While there is a considerable amount of literature on both EFGs and POSGs, the communities using these models do not see much interaction and sharing of results. Therefore, besides slightly augmenting POSGs, we thoroughly analyze the differences and the similarities between the models.
We show that we do not need to view POSGs and EFGs as competing models --- instead, a POSG can be viewed as the \textit{underlying} model from which one \textit{derives} the EFG representation.
Since many research questions depend only on those aspects of the game that are preserved by the EFG representation, most of the existing EFG results are directly applicable to POSGs.
Moreover, the relationship between POSGs and EFGs provides an opportunity for using novel tools in EFGs, discovering new connections between EFGs and POSGs, and bringing existing results from either community to the attention of a wider audience.
However, to fully utilize this relationship, we first need to extend the POSG model with factorized observations.

\subsection*{Outline and Contributions}

We introduce \modelName{} (\modelAbbrev{}), a generalization of the POSG model that factorizes each player's observations to public and private (Section~\ref{sec:model}).
In Section~\ref{sec:EFR}, we show that by ``unrolling'' a \modelAbbrev{}, we can obtain its tree-shaped \textit{extensive-form representation}. This representation is similar to the standard formalization of EFGs, except for being augmented by the information about public knowledge and about the players' knowledge outside of their turn.
Note that rather than being ``yet another new model'', this formalization of EFGs merely takes several changes to the classical definition that already appeared in various recent works \cite{MCCR, brown2018depth, seitzDLS} and makes them explicit.
While we prefer referring to this formalization simply as ``EFG'', this paper shall \edit{sometimes} use the term ``augmented EFG'' to distinguish \edit{this formalization from} the historical one.

We proceed with analyzing the expressive power of the EFG and \modelAbbrev{} models.
EFGs are slightly more general, because
they enable modelling situations where the environment prevents the agents from tracking time (i.e., non-timeable games \cite{timeability}) or forces them to forget a particular piece of information they previously learned (i.e., imperfect recall \cite{wichardt2008}).
Importantly, no game played in the real world against people can have these properties.
As a result, research in EFGs typically assumes that none of the studied games has these properties (unless the properties are the primary focus of the investigation).
From now on, we will, therefore, refer to games without these properties as ``well-behaved''.

\begin{figure*}[tb]
    \includegraphics[width=0.95\textwidth]{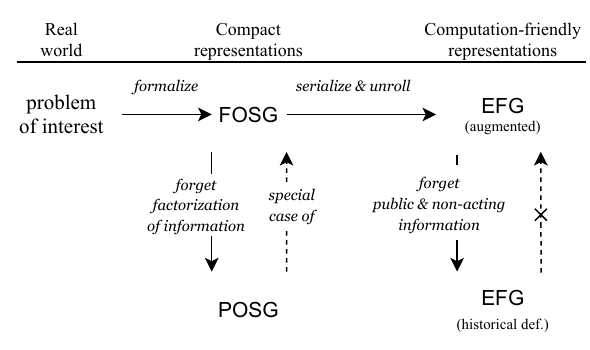}
    \caption{\edit{The relationship between \modelAbbrev{}s, POSGs, and EFGs.
        All timeable perfect-recall EFGs can be obtained by starting with some \modelAbbrev{} (Theorem~\ref{thm:FOSGs-as-EFGs}).
        Going from augmented to ``classical'' EFGs is a lossy process that cannot, in general, be reversed (Section~\ref{sec:EFG-problems}).}
    }
    \label{fig:flow_diagram}
\end{figure*}

\begin{figure*}[tbp]
    \centering
    \includegraphics[width=0.825\textwidth]{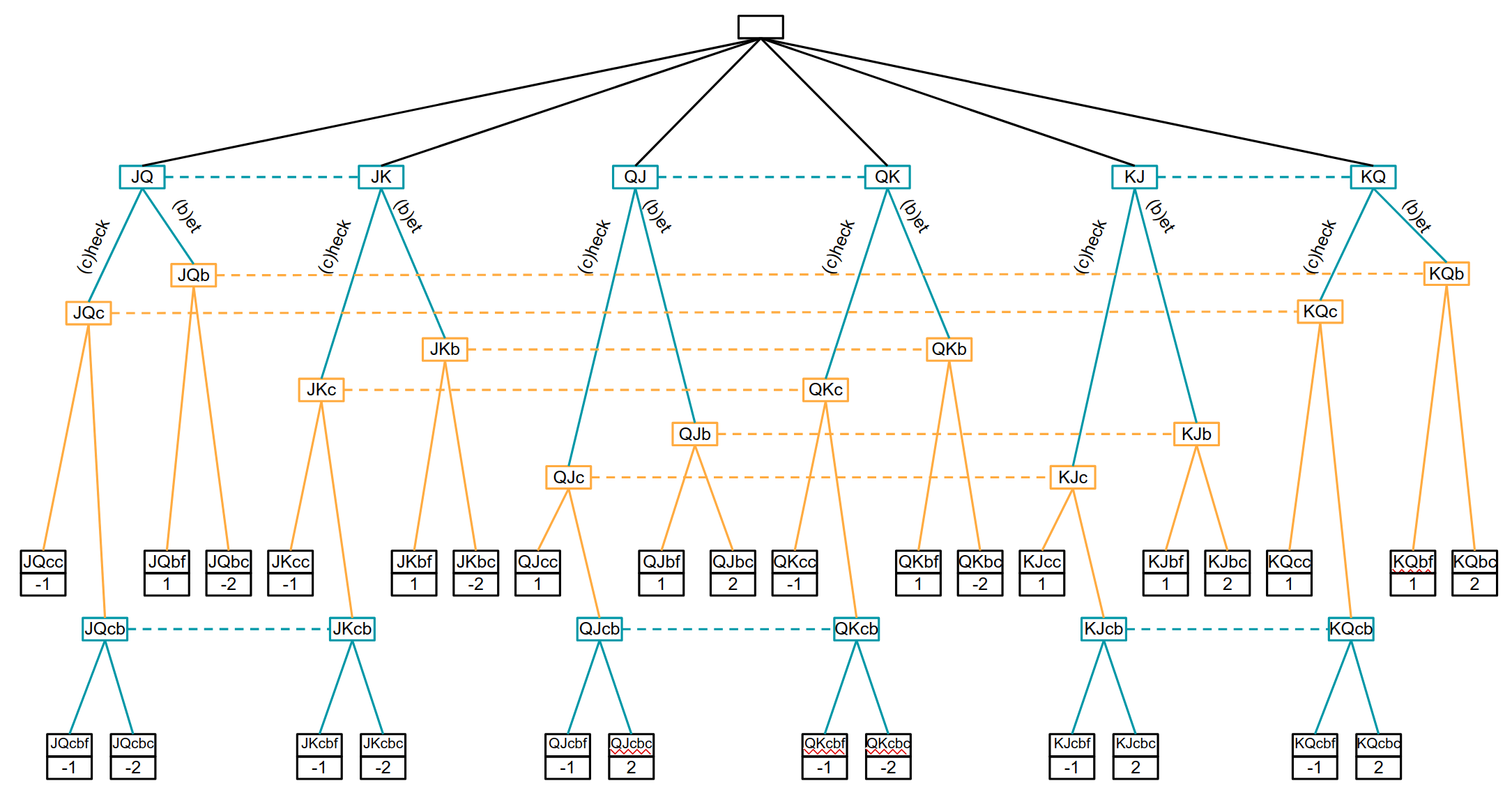}
    \caption{A classical EFG representation of Kuhn poker \edit{(where the only cards are J, Q, and K)}.
    \edit{The dashed lines connect histories in the same information set.}
    There is no notion of public information or information that a player has \edit{outside of their turn}.
    }
    \label{fig:kuhn_efg}
    \centering
    \includegraphics[width=0.825\textwidth]{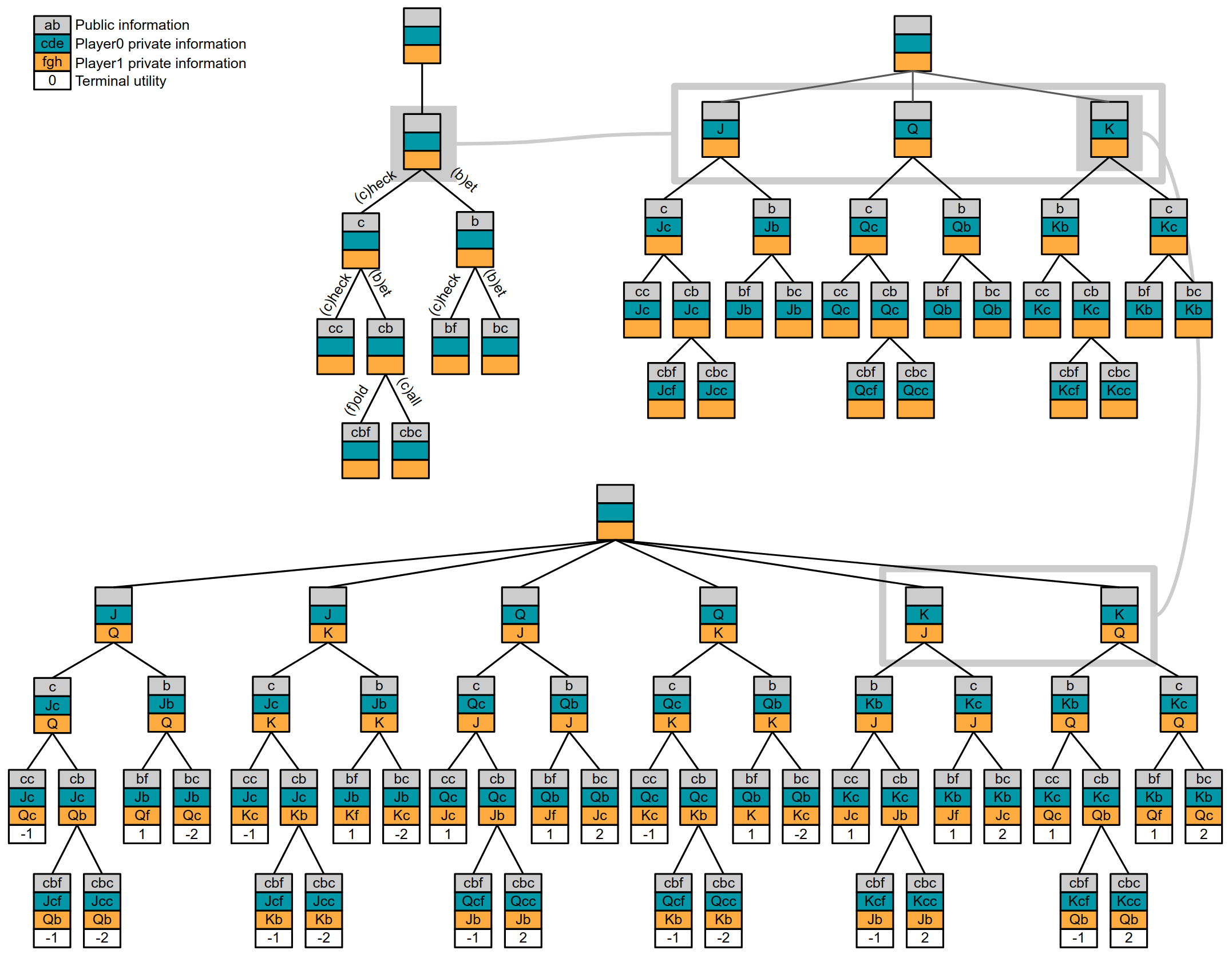}
    \caption{The different points of view \edit{enabled by the} proposed \modelAbbrev{} model.
    Bottom: \edit{the full extensive form representation} of $G$.
    \edit{While the tree structure is as in Figure~\ref{fig:kuhn_efg}, we retain the factorization of information and the information of the non-acting player}.
    Top right: the Player~1 point of view, \edit{obtained by grouping histories based on which action-observation sequence they correspond to}.
    Top left: \edit{the public tree, obtained by grouping action-observation sequences based on which public states they correspond to.}
    }
    \label{fig:kuhn_fog}
\end{figure*}

We show that any well-behaved augmented EFG can be viewed as an extensive-form representation of some \modelAbbrev{} (Theorem~\ref{thm:FOSG-to-aug-EFG}).
To connect the result to the historical ``non-augmented'' formalization of EFGs, we prove that any well-behaved non-augmented EFG can be obtained by stripping the augmented structure from the extensive-form representation of some \modelAbbrev{} (Theorem~\ref{thm:FOSGs-as-EFGs}).
We then explain why this stripping-away process irrecoverably loses some of the information that is necessary for tracking public information in the game (Section~\ref{sec:EFG-problems}).
\edit{For a summary of these relationships, see Figure~\ref{fig:flow_diagram}.}
To build more intuitions about the formalism, the paper includes a graphical illustration of EFGs, \modelAbbrev{}s, and of the latter's better suitability for decomposition (Figures~\ref{fig:kuhn_efg} and \ref{fig:kuhn_fog}).
We also include a ``dictionary'' of key concepts from reinforcement learning, EFGs, and \modelAbbrev{}s (Table~\ref{fig:translatino_table}).

Section~\ref{sec:translating_results} showcases how the model can be used to transfer results between game theory (studied in EFGs) and multiagent reinforcement learning (studied in POSGs).
In particular, Section~\ref{sec:CFR} re-states in \modelAbbrev{}s the counterfactual regret minimization algorithm, a work-horse that serves as a crucial building block of the recent milestones in EFGs.
The \modelAbbrev{} formalism allows to naturally explain this algorithm in terms of state and state-action values, making it particularly accessible to the reinforcement learning community.
Section~\ref{sec:cfr-d} shows that the \modelAbbrev{} framework naturally fits with the notion of subgames, decomposition, and subgame solving --- key building blocks of the search-based methods that brought recent breakthroughs in large imperfect-information games.
Section~\ref{sec:app:sequence_form} then shows how to apply sequence form -- another key technique in developed for EFGs -- to \modelAbbrev{}s.
\edit{Finally, Section~\ref{sec:disc} reviews the most-related work and presents the main takeaways.}
The appendix contains the proofs of the presented results.

    \begin{table*}[tbp]
    \centering
    \footnotesize{
    \begin{tabular}{| r l | r l | r l |}
    \multicolumn{2}{ c }{RL} & \multicolumn{2}{ c }{EFG} & \multicolumn{2}{c }{\modelAbbrev{}}\\ \hline  \multicolumn{6}{ | c | }{ \cellcolor[HTML]{C0C0C0} The actor in the task.} \\ \hline
    $i$ & agent                        & $i$ & player            & $i$ & agent, player           \\
    \multicolumn{6}{ | c | }{\cellcolor[HTML]{C0C0C0} The acting rule.} \\ \hline
    $\pi$ & policy                  & $\sigma \in \Sigma$  & strategy & $\pi$ / $\sigma$ & policy/strategy \\
    \multicolumn{6}{ | c | }{\cellcolor[HTML]{C0C0C0} A measure of how long the task takes.} \\ \hline
    $T$ & finite horizon                  & $D$  & game depth & $T$ & finite horizon \\
    \multicolumn{6}{ | c | }{\cellcolor[HTML]{C0C0C0} The basic notion for describing the ``state of the world''.} \\ \hline
    $s \in \mc S$ & state              & $h\in \mc H$ & history & $w\in \mc W$ & world state             \\ 
    \multicolumn{6}{ | c | }{\cellcolor[HTML]{C0C0C0} A list of everything that has happened so far, from the objective point of view.} \\ \hline
    -- & trajectory    & $h \in \mc H$ & history      & \shortstack{$h\in \mc H$\\$t$} & \shortstack{history\ \ \ \ \ \\trajectory}         \\
    \multicolumn{6}{ | c | }{\cellcolor[HTML]{C0C0C0}Information based on which actors select their next actions.} \\ \hline
    \shortstack{ $s$ \\ $b$} & \shortstack{state (MDP) \ \ \ \ \\ belief (POMDP)} & $I\in \mc I_i$ & infoset   & $s_i \in \mc S_i$ & \shortstack{action-obs.\\history\ \ \ \ \ \ } \\
    \multicolumn{6}{ | c | }{\cellcolor[HTML]{C0C0C0}The immediate feedback on how well the actor is currently doing.} \\ \hline
    $r$ & reward &  & doesn't exist & $\mc O_i(\cdot)$ & observation     \\  
    \multicolumn{6}{ | c | }{\cellcolor[HTML]{C0C0C0}The quantity that actors optimize for.} \\ \hline

     &  cum. reward           & $u_i$ & utility              & \shortstack{$u_i(h)$\\$\mc R_i(h)$} &  \shortstack{utility\ \ \ \ \ \ \ \ \ \ \\cum. reward}  \\
    \multicolumn{6}{ | c | }{\cellcolor[HTML]{C0C0C0}The desirability of a given situation for an actor who uses a given acting rule.} \\ \hline
     $V^\pi(s)$ & $s$'s value for $\pi$ & -- & exp. utility at $h$ & $v^\sigma_i(h)$ & history value \\
    \multicolumn{6}{ | c | }{\cellcolor[HTML]{C0C0C0}The cause of non-determinism in the task.} \\ \hline
    $\mc T(s,\! a)$ & transition prob. & $\sigma_c(h,a)$ & chance strategy & $\mc T(w,a)$ & transition prob. \\
    \multicolumn{6}{ | c | }{\cellcolor[HTML]{C0C0C0}The probability of ending up in a given situation.} \\ \hline
     &  & $\pi^\sigma(h)$ & reach probability  & $P^\pi(h)$ & reach prob.\\
    \hline
    \end{tabular}
    }
    \vspace{1em}
    \caption{A ``dictionary'' of the basic concepts of MDPs (and other models used in RL), EFGs, and \modelAbbrev{}s. The correspondences are on the level of usage; for example, the correspondence between a state in an MDP and a history in an EFG does not imply they are mathematically equivalent, but rather that the role states play in MDPs is analogous to the role histories play in EFGs.}
    \label{fig:translatino_table}
\end{table*}
\section{\ModelNameCamel{}s}\label{sec:model}
    In this section, we describe the \modelName{} model as a variation on partially-observable stochastic games.
Since a key feature of the model is its ability to talk about public information, let us first explain what this concept refers to, why is it important, and why incorporating it into our models is natural, useful, and inexpensive.

\subsection{Public Information and Decomposition}\label{sec:motivation:public}

A piece of information is said to be \defword{common knowledge} among a group of agents if all the agents know it, they all know that they know it, they all know that they all know that they know it, and so on \cite{fagin2003reasoning}.
Figuring out which information is common knowledge is often difficult, as it requires putting oneself in shoes of the other agents and accurately reasoning about their thought processes. (The Wikipedia page on common knowledge has a puzzle that illustrates this well \cite{comm_know_wiki}.)
In contrast, a piece of information is \defword{public} among a group of agents if each agent received it in a way that trivially reveals that the information is common knowledge.
Some actions that create public knowledge are (a) speaking out loud in a group (b) placing a card face-up, or (c) moving a piece on a game board.

To see why public information is important, first note that
the essence of game theory is that playing well requires knowing which actions the other players might take, which tends to involve reasoning about the information they have.
Knowing that some information $I$ is \textit{common knowledge} is therefore immensely useful for decomposition.
Indeed, each player will only reason about situations compatible with $I$ (and about others' reasoning about situations compatible with $I$, others' reasoning about reasoning about situations compatible with $I$, etc.).
Situations compatible with $I$ can, therefore, be considered mostly independently of those incompatible with it (the limited dependence is explained in, e.g., \cite{DeepStack}).
However, finding out which information is common knowledge can be costly, so the best approach will often be to decompose games based on the subset of common knowledge that is \textit{public}.
Once we can decompose problems into smaller pieces, we become able to solve large problems that would otherwise be intractable.
Indeed, we have recently seen breakthroughs in a number of problems that were made computationally feasible by decomposition based on public knowledge ---
examples include poker \cite{DeepStack,Libratus,Pluribus}, Hanabi \cite{foerster2019bayesian,lerer2020improving}, general EFGs \cite{MCCR,li2020regret,davis2019low}, Dec-POMDPs \cite{spaan2008multiagent,de2019multi}, and one-sided POSGs \cite{horak_public}.

However, the existing game-theoretic and MARL models do not have a built-in way of describing public knowledge.
This is a serious problem because figuring out whether a piece of information is public or not might be difficult (or impossible) unless one remembers how the information was obtained.
(Indeed, suppose I know that a friend would be happy to pick me up from an airport, and they know I would be glad if they did. This alone says nothing about whether the information is public among us, so unless we talked about this, I should probably take a taxi. For more technical examples, see Section~\ref{sec:EFG-problems}.)
Making use of decomposition thus typically required adding some component on top of the employed formal model.
In addition to requiring a non-trivial conceptual and technical effort, this was often done in a domain-specific manner \cite{DeepStack,Pluribus}, which made the devised methods difficult to generalize.
If the generalization were straightforward, the situation in imperfect-information games would by now likely be closer to that in perfect-information games or single-agent RL, where many of the state-of-the-art algorithms are very general \cite{alphaZero,muZero,agent57}.
This suggests that using a model which keeps track of public information by default would have significant benefits.

Fortunately, as witnessed by the examples (a), (b), and (c), the information about which knowledge is public is inherently a part of the description of many games and real-world situations.
In other words, public knowledge is typically \textit{already} a part of a problem's natural definition; the model we propose merely \textit{preserves} this information while \textit{models used in the past discard it}. 
    \subsection{Description of the Basic Model}\label{sec:model:def}
        Informally, the model we are about to describe captures a situation where multiple actors take actions -- possibly simultaneously -- which influence how the world's state changes.
The new world-state might be more or less desirable for individual actors, which is captured by corresponding reward functions.
Rather than having full knowledge of the world state, the agents perceive it through observations.
These are further ``factored'', such that if some information is public, each agent will know that everybody else also has access to it.
The various footnotes in this section explain the motivation behind the presented definitions and suggest possible extensions of the model. These might be helpful for some readers, but can be otherwise skipped.
In the following paragraph, $\hookrightarrow$ indicates a \textit{partial function}, defined on a subset of the stated domain.

A \defword{\modelName{}} (\modelAbbrev) is a tuple $G = \left< \mc N, \mc W, p, \initState, \mc A, \mc T, \mc R, \mc O \right>$, where
$\mc N$ is the \defword{player set},
$\mc W$ is the set of \defword{world states},
$\initState$ is a designated \defword{initial state},
$p : \mc W \to 2^{\mc N}$ is a \defword{player function},
$\mc A$ is the space of \defword{joint actions},
$\mc T : \mc W \times \mc A \hookrightarrow \Delta \mc W$ is the \defword{transition function},
$\mc R : \mc W \times \mc A \hookrightarrow \R^{\mc N}$ is the \defword{reward function},
$\mc O : \mc W \times \mc A \times \mc W \hookrightarrow \mb O$ is the \defword{observation function}\footnote{While deterministic observations are the norm in the EFG literature and much of MARL, researchers from fields like robotics can use the straightforward stochastic modification of this setting where $\mc O$ maps into $\Delta \mb O$ instead. This is equivalent to the standard setting, since one can always emulate stochastic observations by adding a hidden variable into $w$.},
and we have:

\medskip

\begin{itemize}
    \item $\mc N = \{1,\dots,N\}$ for some $N\in \mb N$.
    \item $\mc W$ is compact. For formal convenience\footnote{An initial state with no player allows us to both emulate any initial belief over $\mc W$ and assume that each player's information state starts with an observation (which essentially states ``Game started. Legal actions = $\{ \ldots \}$.'' and is helpful for model-free agents.).}, we assume that $p(\initState) = \emptyset$.
    \item $\mc A = \prod_{i \in \mc N} \mc A_i$, where each $\mc A_i$ is an arbitrary set of \defword{$i$'s actions}.
        \begin{itemize}
            \item For each $i\in p(w)$, $\mc A_i(w) \subset \mc A_i$ denotes a non-empty compact set of $i$'s (legal) \defword{actions at $w$}.
                We denote $\mc A(w) := \prod_{i \in p(w)} \mc A_i(w)$.
            \item We denote $\mc A_i(w) := \{noop\}$ for $i\notin p(w)$, which allows us to identify each $a \in \mc A(w)$ with an element of $\prod_{i \in \mc N} A_i(w)$ by appending to it the appropriate number of \textit{noop}\footnote{\edit{Recall that \textit{noop} stands for ``no operation'' ---  an action that makes no changes.}} actions.
        \end{itemize}
    \item The transition probabilities $\mc T(w,a) \in \Delta \mc W$, $a\in \mc A(w)$, are defined for all $w\in \mc W$ with non-empty $p(w)$ and for some $w$ with no active players.
        \begin{itemize}
            \item A world state with $p(w)=\emptyset$ and undefined $\mc T(w,a)$ is called \defword{terminal}.
        \end{itemize}
    \item $\mc R (w,a) = (\mc R_i(w,a))_{i \in \mc N}$ for each non-terminal state $w$ and $a\in \mc A(w)$.\footnote{As is common in MDPs, we could equivalently consider rewards that depend on $(w,a,w')$, or just on $w'$.}
    \item $\mc O$ is factored into \defword{private observations} and \defword{public observations} as $\mc O = (\mc O_{\textnormal{priv}(1)},\dots,\mc O_{\textnormal{priv}(N)}, \mc O_{\textnormal{pub}} )$.
    \begin{itemize}
        \item $\mb O = \prod_{i\in\mc N} \mb O_{\textnormal{priv}(i)} \times \mb O_{\textnormal{pub}}$\edit{, where $\mb O_{(\cdot)}$ are arbitrary sets (of possible observations)}.
        \item We assume that $\mc O_{(\cdot)}(w,a,w') \in \mb O_{(\cdot)}$ is defined for every non-terminal $w$, $a \in \mc A(w)$, and $w'$ from the support\footnote{For finite $\mc W$, being in support of $\mc T(w,a)$ is equivalent to having a non-zero probability.} of $\mc T(w,a)$.
    \end{itemize}
\end{itemize}

\bigskip

The game proceeds as follows:
It starts in the initial state $\initState$.
In each state, each active player $i \in p(w)$ learns which actions are legal for them (either by deduction or by being explicitly told) and selects $a_i \in \mc A_i(w)$.
(While the player obviously knows which action \textit{they} took, they often will not know the actions of the other players, or possibly not even which of them were active at $w$.)
The game then transitions to a new state $w'$, drawn from the distribution $\mc T(w,a)$ that corresponds to the joint action $a=(a_i)_{i\in p(w)}$.
This generates the observation $\mc O(w,a,w')$, from which each player receives $\mc O_i(w,a,w') := \left( \mc O_{\textnormal{priv}(i)}(w,a,w'), {\mc O}_{\textnormal{pub}} (w,a,w') \right) \in \mb O_i$ (i.e., the public observation together with their private observation, in manner which allows distinguishing between the two).\footnote{There are games -- such as Hanabi -- where information might be public among a \textit{subset} of players. Such decomposition can be modelled by factoring $\mc O$ into $(\mc O_P)_{P \subset \mc N}$, with each player receiving $\mc O_i = ((P,\mc O_P))_{i \in P \subset \mc N}$. The default \modelAbbrev{} model is a special case of this generalization, with $\mc O_P$ being empty for every $P$ other than either $P=\mc N$ or $P=\{i\}$.}
Finally, each player is assigned the reward $\mc R_i(w,a)$. However, a player might not know how much reward they received, unless this is -- explicitly or implicitly -- a part of $\mc O_i$.\footnote{For example, a thief might steal a wallet without immediately inspecting its contents, or a robot might be optimizing for my preferences while having uncertainty over them \cite{humanCompatible}. Unobservable rewards also remove incentives to ``hide from bad news'' or ``wirehead'' \cite{everitt2016avoiding}.}
This process repeats until a terminal state is reached.
The goal of each player is to maximize the sum of rewards $\mc R_i(w,a,w')$ obtained during the game.
Finally, Remark~\ref{rem:chance_player} describes an important extension of the basic \modelAbbrev{} model:

\begin{remark}[Chance player]\label{rem:chance_player}
In many domains, it is useful to view the stochasticity in the game as being caused by a \defword{chance player} $c$ (also called \defword{nature}).
The chance player receives observations and takes actions like any other player, but their policy (formally defined in Section~\ref{sec:model:derived}) is fixed and publicly known.
The addition of $c$ causes no loss of generality since their actions can always be merged back into the transition function $\mc T$.
\end{remark}
        \edit{We illustrate the above definitions on the example of Kuhn poker:}

\begin{example}[Kuhn poker as a \modelAbbrev{}]\label{ex:kuhn}
\edit{Kuhn poker is a form of poker where the deck includes only three cards -- Jack, Queen, and King.
First, each player places one chip into the \emph{pot} as the initial ``forced'' bet (\emph{ante}).
Each player is then privately dealt one card (the last card isn't revealed).
This is followed by a betting phase (explained below).
The game ends either when one player \emph{folds} (forfeiting all bets made so far to their opponent) or there is a \emph{showdown}, where the private cards are revealed and the higher card's owner receives the bets.
At the start of the betting, player one can either \emph{check} (bet no chips) or \emph{bet} (one chip).
If they \emph{check}, player two can also \emph{check} --- leading to a \emph{showdown} --- or \emph{bet}.
If one of the players \emph{bets}, their opponent must either \emph{call} (betting one chip to match the opponent's bet), followed by a \emph{showdown}, or \emph{fold}.}

\edit{One way of modelling the game\footnote{\edit{There will typically be multiple ways of modelling a game. For example, we could omit the (strategically trivial) \emph{ante} actions, deal both private cards at once, or end the game immediately after the \emph{call} actions. Such more compact model is depicted in Figure~\ref{fig:kuhn_fog}.}} as a \modelAbbrev{} is to view world states as tuples
$w = (C_1, C_2, \textnormal{chips}_1, \textnormal{chips}_2, \textnormal{pl})$,
where
$C_i \in \{\emptyset\} \cup \textnormal{Deck}$, $\textnormal{Deck} := \{ J, Q, K \}$,
$\textnormal{chips}_i \in \{0, 1, 2 \}$,
and $\textnormal{pl} \in \{ \emptyset, 1, 2, \textnormal{c} \}$
correspond to the private cards, remaining chips, and the currently-active player.
Ignoring the \textit{noop} actions for brevity and
denoting $w' = (\dots, \, \textnormal{chips}'_i, \, \dots)$,
we set
$p(w) := \textnormal{pl}$,
$\mc A_i := \{ \textnormal{ante, check, bet, call, fold} \}$,
and $\mc R_i(w,a,w') := \textnormal{chips}'_i - \textnormal{chips}_i$.
Using $C$, $? \to i$, and $(C_1, C_2)$ as formal shorthands for ``your private card is $C$'', ``unknown card dealt to player $i$'', and  ``showdown: the private cards were $C_1$ and $C_2$'',
the observation spaces can be modelled as
$\mb O_{\textnormal{priv}(i)} := \{\emptyset\} \cup \textnormal{Deck}$ and
$\mb O_\textnormal{pub} := \mc A_1 \cup \{ ?\to 1, ? \to 2 \} \cup \textnormal{Deck}^2$.}
\edit{To point at the formal definition of $\mc A_i(w)$, $\mc T$, and $\mc O$, we look at one trajectory in detail.
The game always starts in $w^0 = w^{\textnormal{init}} = ( \emptyset, \emptyset, 2, 2, 1 )$ with $\mc A_1(w^\textnormal{init}) = \{\textnormal{ante} \}$,
from which it transitions to $w^1 = ( \emptyset, \emptyset, 1, 2, 2 )$.
(Formally, $\mc T(w^0, \textnormal{ante}, w)$ is $1$ when $w = w^1$ and $0$ otherwise.)
In $w^1$, we have $\mc A_2(w^1) = \{ \textnormal{ante} \}$
and the game transitions to $w^2 = ( \emptyset, \emptyset, 1, 1, \textnormal{c} )$.
Private cards are dealt next, via a uniformly random transition; first to $w^3 = (C_1, \emptyset, 1, 1, \textnormal{c} )$, $C_1 \in \textnormal{Deck}$,
and then to $w^4 = (C_1, C_2, 1, 1, 1)$, $C_2 \in \textnormal{Deck} \setminus \{C_1\}$.
Suppose that $C_1 = Q$ and $C_2 = K$.
We have $\mc A_1(w^4) = \{ \textnormal{check, bet} \}$ and if player one \emph{bets}, the game transitions to $w^5 = (Q, K, 0, 1, 2)$, $\mc A_2(w^5) = \{ \textnormal{call, fold} \}$.
If player two responds by \emph{calling}, the game transitions first to $w^6 = (Q, K, 0, 0, \textnormal{c})$ and then to the terminal state $w^7 = (Q, K, 0, 4, \emptyset)$ (since $K$ is higher than $Q$).
The corresponding observations $\mc O_1 (w^{k-1}, \, \cdot \, , w^k)$ are:
1) $(\emptyset, \textnormal{ante})$
2) $(\emptyset, \textnormal{ante})$
3) $(Q, ? \to 1)$
4) $(\emptyset, ? \to 2)$
5) $(\emptyset, \textnormal{bet})$
6) $(\emptyset, \textnormal{call})$
7) $(\emptyset, (Q, K) )$.
}
\end{example}
    \subsection{\modelAbbrev{}s as an Extension of POSGs}\label{sec:POSGs}
        There are many formalizations and variants of POSGs in the literature \citep{peshkin2000learning, POSGs2004hansen,POSGsCommPayoff2004emery,horak_public}.
However, they all behave similarly to the following definition:

\begin{definition}[POSG]\label{def:POSG}
A \defword{partially-observable stochastic game} is a \modelAbbrev{} in which
$\mc O_{\textnormal{pub}}$ is constantly equal to $\emptyset$ and
$p(w)=\mc N$ for all $\initState \neq w \in \mc W$.
\end{definition}

For the purposes of this text, we can thus view POSGs as a special case of \modelAbbrev{}s (i.e., those in which the player function and all public observations are trivial).
In particular, this shows that \modelAbbrev{}s can be viewed as a minor extension of POSGs.

The relationship is even stronger since forgetting the factorization of observations yields a POSG that is strategically equivalent to the original \modelAbbrev{} (i.e., each player has the same information as before, makes the same choices, and these lead to the same outcomes).
Indeed, for a \modelAbbrev{} $G = \left< \mc N, \mc W, p, \initState, \mc A, \mc T, \mc R, \mc O \right>$
we can denote by $\texttt{ForgetFactorization}(G)$ the \modelAbbrev{} $\left< \mc N, \mc W, p', \initState, \mc A, \mc T, \mc R, \mc O' \right>$,
where $p'(w) := \mc N$ for $w\neq \initState$,
$\mc O'_{\textnormal{priv}(i)} := \mc O_i$,
and $\mc O'_{\textnormal{pub}}(\cdot) = \emptyset$.
We immediately get the following:

\begin{proposition}\label{prop:POSG-ization}
For every \modelAbbrev{} $G$, $G' = \texttt{ForgetFactorization}(G)$ is a partially-observable stochastic game.
\end{proposition}

Nevertheless, the resulting POSG will be different in one potentially important aspect:
to reason about public information, we will first need to recover the factorization that we chose to forget in the first place.
This implies that the POSG and \modelAbbrev{} frameworks are formally equivalent, and any research done in one will transfer to the other.
However, the latter will be more convenient for methods which can take advantage of decomposition.

    \subsection{Histories, Information States, and Public States}\label{sec:model:derived}

With access to the underlying FOSG model, we can consider several derived objects which allow for its more nuanced analysis.

To describe the real state of affairs in $G$, the basic concept is that of a \defword{trajectory} --- a finite sequence $\tau = w^0 a^0 w^1 a^1 \dots w^t \in (\mc W \mc A )^* \mc W$ for which $a^k \in \mc A(w^k)$, $w^{k+1} \in \textnormal{supp}\, \mc T(w^k,a^k)$ --- and $i$'s \defword{cumulative reward} $\mc R_i(\tau) := \sum_{k=0}^{t-1} \mc R_i(w^k,a^k)$ along $\tau$.
By $\mc H$ we denote the set of all \defword{histories} in $G$, i.e., all trajectories that start in the initial state $\initState$.
Using the symbol $\sqsupset$ to denote the fact that one sequence extends another, we endow $\mc H$ with the structure of a tree.
Using the set of terminal histories\footnote{Since the last state in each $\tau$ uniquely defined, the notation for world-states can be overloaded to work with trajectories and histories as well. That is, $h= w^0\ldots w^k$ is said to be terminal if $w^k$ is, we set $\mc A(h) := \mc A(w^k)$, etc.} $\mc Z := \{ z \in \mc H \mid z \textnormal{ is terminal} \}$, we define $i$'s \defword{utility} function $u_i : \mc Z \to \R$ as $u_i(z) := \mc R_i(z)$.

To describe the point of view of an individual agent $i$, we talk about \defword{information states} $s_i$.
To simplify the notation, we assume that $i$ can always deduce which actions are legal from other information available to them.
Initially, each player's information state is an empty sequence $s_i = \emptyset$.
When $i$ receives an observation $O_i = \mc O_i(w,a,w')$, resp. takes a non-\textit{noop} action $a_i$, $s_i$ changes to $s'_i = s_i O_i$, resp. $s'_i = s_i a_i$.
In light of this definition, information states can also be called \defword{action-observation histories}.
Each history corresponds to some information state $s_i(h)$ that ends by an observation and, by our assumption, we have $\mc A_i(h) = \mc A_i(g) =: \mc A_i(s_i)$ whenever $s_i(h) = s_i(g) = s_i$.
When defining \modelAbbrev{}s, we have assumed that whenever the world-state changes, \textit{all} players receive some observation (i.e., $\mc O_i(w,a,w')$ is defined for all legal transitions).
This is equivalent to allowing the players to write even \textit{noop} actions into their $s_i$-s and enables them to always deduce the number of transitions that have occurred so far.
This assumption has non-trivial -- but mostly positive -- consequences, as we will see in Theorem~\ref{thm:FOSGs-as-EFGs}.\footnote{Arguably, it would be more natural to allow observation functions which are only defined on a \textit{subset} of legal transitions and assume that players for whom $\mc O_i(w,a,w')$ is undefined are oblivious to the fact that the transition has happened. Unfortunately, this makes all EFG-related concepts more complicated (by introducing ``thick infosets'' \cite{seitzDLS} and ``non-timeability'' \cite{timeability}), so we have decided to not include this option in the default \modelAbbrev{} model.}

The \defword{information-state tree} of $i$ is the set $\mc S_i$ of all $s_i$-s that might arise in $G$.
The \defword{behavioural strategy} (also called \defword{policy}) of $i$ is a mapping $\pi_i : \mc S_i \hookrightarrow \Delta \mc A_i$ s.t. $\pi_i(s_i) \in \Delta \mc A_i(s_i)$ for every $s_i$ where $i$ is supposed to submit an action.
Any \defword{strategy profile} $\pi = (\pi_i)_{i\in \mc N}$ induces a probability distribution over terminal histories.
We will henceforth assume that the game is guaranteed to always end up in a terminal state\footnote{This can be ensured by assuming that $\mc W$ is finite and $\mc T$ admits no loops or that $\mc T$ makes infinite loops or chains vanishingly unlikely, or by enforcing a finite horizon $T$.},
which allows us to define the expected utility $u_i(\pi) := \mathbf E_{z \sim \pi} \, u_i(z)$ and talk about various solution concepts in \modelAbbrev{}s (such as Nash equilibria, in which no player can increase their expected utility by deviating unilaterally from the current strategy profile).\footnote{\edit{Note that the computational requirements of many algorithms (e.g., CFR \cite{CFR}) will scale with the size of $\mc S_i$, which can be exponentially larger than the size of $\mc W$.
Indeed, many algorithms reason about policies and each policy $\pi_i$ specifies a probability distribution over $\mc A_i$ \textit{for every information state in $\mc S_i$.}
To see that $| \mc S_i |$ is potentially much larger than $| \mc W |$, note that (a) trajectories in $G$ can visit states multiple times, so there are potentially (exponentially) more trajectories than world states and (b) depending on how much information about $G$ a player has, they can have up to as many infostates as there are trajectories.}}

To describe the point of view of an ``external observer'', it is useful to consider the notion of \defword{public states}.
Formally, a public state $\pub = \pub(h)$ corresponding to a history $h$ is a sequence of public observation received along $h$.
By $\pubTree$, we denote the \defword{public tree} which consists of all public states that can arise in $G$.

Note that since each player has their observations factored into $\mc O_{\textnormal{priv}(i)}$ and $\mc O_{\textnormal{pub}}$, we can equally-well recover $\pub$ from any $s_i = s_i(h)$.
And while this fact might seem essentially trivial, it makes for a very powerful decomposition tool in \modelAbbrev{}s.
Indeed, if the current public state is $\pub$, each player knows that \textit{all other players} can condition their strategy on $\pub$, so each player can restrict their reasoning to situations that are compatible with $\pub$.
Decomposition in \modelAbbrev{}s is further discussed in Section~\ref{sec:cfr-d}.

\edit{For example, the first two transitions described in Example~\ref{ex:kuhn} generate the history
$h = \left( (\emptyset, \emptyset, 2, 2, 1) , \textnormal{ante} , (\emptyset, \emptyset, 1, 2, 2) , \textnormal{ante} , (\emptyset, \emptyset, 1, 1, \textnormal{c}) \right)$.
As this is unwieldly for long histories, we can identify each such $h'$ with a shorter sequence of events that uniquely identify it.
For example, the history}
\vspace{-0.5em}
\[
h' = (w^0, \textnormal{ante}, \dots, \textnormal{bet}, w^5) = ( (\emptyset, \emptyset, 2, 2, 1), \textnormal{ante}, \dots , (Q, K, 0, 1, 2) )
\]
(where $w^k$ are as in Example~\ref{ex:kuhn}) could be more succinctly written as
$h' = (Q, K, \textnormal{bet} )$.
The corresponding infostate and public state are
\vspace{-0.5em}
\begin{equation*}
s_1(h') = (\textnormal{ante}, (\emptyset, \textnormal{ante}), (\emptyset, \textnormal{ante}), (Q, ?\to 1), (\emptyset, ? \to 2), \textnormal{call}, ( \emptyset, \textnormal{bet}) )
\end{equation*}
and 
$\pub = ( \textnormal{ante}, \textnormal{ante}, ?\to 1, ? \to 2, \textnormal{bet} )$,
which can be identified with just 
$( Q, \textnormal{bet} )$ and $(\textnormal{bet})$.\footnote{\edit{Note that omitting the observation $?\to 2$ \emph{before taking the \emph{bet} action} would make indistinguishable the situations where player two already has, resp. doesn't have, their private card.}}
\edit{For a visual illustration, see Figure~\ref{fig:kuhn_fog}.}
\section{Extensive Form Representation of FOSGs}\label{sec:EFR}
	In this section, we describe and analyze extensive-form representations of \modelAbbrev{}s.
Informally, such representations are constructed by ``unrolling'' the possibly-cyclic space $\mc W$ into the tree-structured set $\mc H$.
This makes them less compact than the default model but, as demonstrated in Section~\ref{sec:translating_results}, more suitable for various modern algorithms.

In Section~\ref{sec:sub:efr}, we start with an arbitrary \modelAbbrev{} and construct its extensive-form representation.
This construction does not assume prior knowledge of EFGs --- in fact, one goal of this section is to show that even if EFGs did not yet exist, they would naturally arise as an object derived from \modelAbbrev{}s.
(Readers familiar with the EFG literature will notice that the resulting structure is subtly yet crucially different from the historical formalization of EFGs. However, rather than ``introducing yet another new model'', the proposed definition essentially amounts to gathering in one place several modifications that were already present in several existing publications.)
We also show that EFGs can be considered even in the absence of the underlying \modelAbbrev{} and prove that while this might introduce certain complications, this never happens with EFGs derived from \modelAbbrev{}s.
In Section~\ref{sec:sub:FOSGs-as-EFGs},
we compare the proposed formalization of EFGs with the historical one.
Formally, the two definitions are near-identical, except that the latter lacks the means to describe public information and information available to players outside of their turn.
The upshot is that if we start with extensive-form representations of FOSGs and throw the given information away, we obtain precisely all timeable perfect-recall EFGs.
(Both of these limitations are desirable: while we explain that imperfect recall can be added back if needed, \cite{timeability} argues that non-timeable games are pathological and of no practical interest).
In Section~\ref{sec:EFG-problems},
we demonstrate why the information thrown away by classical EFGs is fundamentally irrecoverable and explain some of the resulting the drawbacks.

\subsection{Extensive Form Representation of \modelAbbrev{}s}\label{sec:sub:efr}

In this section, we project the structure of information- and public-states onto the history tree $\mc H$, which allows us to discard the original \modelAbbrev{} and only focus on its tree representation.
However, before doing so,
%
%
we show that every \modelAbbrev{} has a version in which only one player acts at any given time\footnote{Technically speaking, this step isn't necessary for any of the theory. However, it seems useful purely on the grounds of making the resulting representations more similar to classical EFGs.}:

\begin{definition}[Serial \modelAbbrev{}]
A \modelAbbrev{} $G$ is said to be \defword{serial} if for each $w\in \mc W$, either $p(w)=\{i\}$ and all transitions $\mc T(w,a)$ are deterministic or $p(w) = \emptyset$.
A non-terminal $w$ with $p(w) = \emptyset$ is called a \defword{chance node}.
\end{definition}

The serialized version of $G$, denoted $G' := \texttt{Serialize}(G)$, can be constructed by having all active players select actions one by one --- formally transitioning to a new state, but not generating any non-trivial rewards or observations --- and then adding a chance node which determines the next world-state and generates rewards and observations.
The construction is straightforward, results in a strategically equivalent \modelAbbrev{}, and can be viewed as canonical (up to the order of players).
It is formally described in the appendix (Example~\ref{ex:serialization}).

\begin{lemma}\label{lem:serialization}
For every \modelAbbrev{} $G$, the \modelAbbrev{} $G' = \texttt{Serialize}(G)$ is serial.
\end{lemma}


Information in \modelAbbrev{}s is primarily expressed in terms of information \textit{states} which ``live in the heads of agents''.
However, information can also be modelled using information \textit{sets} which ``live'' in the history tree $\mc H$:

\begin{definition}[Information sets]\label{def:infoset}
Let $s_i \in \mc S_i$ be a information state in $G$.
An \defword{information set} (\defword{infoset}) corresponding to $s_i$ is defined as $I_i(s_i) := \{ h\in \mc H \mid s_i(h) = s_i \}$.
The collection $\mc I_i$ of all $i$'s non-empty\footnote{There are information states of $i$ --- those that end in an action --- that are not associated with any particular history, and thus correspond to an empty information set.} infosets is called $i$'s \defword{information partition}.
\end{definition}

\noindent
\edit{For example, in Kuhn poker, the infostate $s_1(Q, ?\to 2, \textnormal{bet})$ corresponds to the infoset $I_1 = \{ (Q,J,\textnormal{bet}), (Q,K,\textnormal{bet}\}$.}

Since the space of information-\textit{states} $\mc S_i$ is endowed with the tree structure given by the extension relation $s_i \sqsubset t_i$, we can likewise turn $\mc I_i$ into a tree by saying that an infoset $J \in \mc I_i$ is an \defword{extension} of $I\in \mc I_i$ if we have $J=I_i(t_i)$ and $I=I_i(s_i)$ for some $s_i \sqsubset t_i$.\footnote{Since $i$ is assumed to have perfect recall, this is equivalent to each history from $J$ being an extension of some history from $I$, which is further equivalent to the existence of a single history from $J$ that is an extension of some history from $I$.}
The \textit{tree} of public states can be defined analogously:

\begin{definition}[Public sets]\label{def:public_set}
Let $\pub \in \pubTree$ be a public state in $G$.
A \defword{public set} corresponding to $\pub$ is defined as $\pubSet(\pub) := \{ h\in \mc H \mid \pub (h) = \pub \}$.
The collection $\pubSetTree$ of all all non-empty public sets is called the \defword{public partition}.
\end{definition}

Extensive form games are obtained by abstracting away the original \modelAbbrev{} and only looking at the corresponding trees of histories, infosets, and public sets:

\begin{definition}[EFG]\label{def:augmented-EFG}
An (augmented) \defword{extensive form game}\footnote{Strictly speaking, this definition is novel. However, explicit attempts to augment $\mc I_i$-s to the full $\mc H$ already appear in \cite{CFR-D}. The first definition of public sets in the CFR literature comes from \cite{accelerated_BR}.} is a tuple $E = \left< \mc N, \mc A, \mc H, p, \pi_c, \mc I, u \right>$ for which
\begin{itemize}
 \item $\mc N = \{1,\dots,N\}$ for some $N \in \mc N$,
 \item $\mc H$ is a tree\footnote{Recall that a ``tree on $X$'' refers to a subset of $X^*$ that is closed under initial segments.} on $\mc A$,
 \item $\mc A$ and all sets $\mc A(h) := \{ a\in \mc A \mid ha \in \mc H \}$, for $h\in \mc H$, are compact,
 \item $p : \mc H \setminus \mc Z \to \mc N \cup \{c\}$ (where $\mc Z$ denotes the leaves of $\mc H$),
 \item $\pi_c(h) \in \Delta \mc A(h)$ for $p(h)=c$,
 \item $u : \mc Z \to \R^{\mc N}$, and
 \item $\mc I = (\mc I_1,\dots, \mc I_N,\pubSetTree)$ is a collection of partitions of $\mc H$, where each $\mc I_i$
 \begin{itemize}
  \item is a refinement\footnote{Recall that $\mc P$ is a refinement of $\mc Q$ if each $P\in \mc P$ is a subset of exactly one $Q \in \mc Q$.} of $\pubSetTree$ and
  \item provides enough information to identify $i$'s legal actions.\footnote{That is, for every $I\in \mc I_i$, $p(h)$ is either equal to $i$ for all $h\in I$ or for no $h\in I$. If for all, then $\mc A(h)$ doesn't depend on the choice of $h\in I$.}
 \end{itemize}
\end{itemize}
\end{definition}

To avoid various pathologies, we might wish to additionally require perfect recall and ``no thick infosets'':
$E$ is with \defword{perfect recall} if for each $g,h\in I \in \mc I_i$, $i$'s action-infoset histories\footnote{Where $i$'s action-infoset history corresponding to $h$ is the sequence of infosets encountered by $i$ and actions taken by $i$ on the way to $h$.} corresponding to $g$ and $h$ coincide.
$E$ is said to not have \defword{thick public sets} if no element of $\pubSetTree$ (and hence of $\mc I_i$) contains both some $h$ and its strict extension.

Moreover, any \modelAbbrev{} $G$ canonically\footnote{Up to the order of players and some formal details.} corresponds to some \defword{extensive-form representation} $\EF{G}$.
Informally speaking, $\EF{G}$ is obtained by serializing $G$, only keeping the objects that appear in Definition~\ref{def:augmented-EFG}, and formally modifying them until it fits the definition.
(The formal construction is described in the proof of Theorem~\ref{thm:FOSG-to-aug-EFG}.)
Not only does this construction indeed produce an augmented EFG, we even get the two convenient assumptions for free:

\begin{restatable}[\modelAbbrev{}s as (augmented) EFGs]{theorem}{AugEFGize}\label{thm:FOSG-to-aug-EFG}
Every \modelAbbrev{} $G$ corresponds to an (augmented) EFG $E = \EF{G}$ with perfect-recall and no thick public sets.
Moreover, any perfect-recall augmented EFG without thick public sets can be obtained this way.
\end{restatable}

Note that Theorem~\ref{thm:FOSG-to-aug-EFG} does \textit{not} imply that \modelAbbrev{}s are unsuitable for studying imperfect recall abstractions \cite{cermak2017ijcai,CRSWFG}.
Indeed, imperfect recall can be introduced either (1) in the underlying \modelAbbrev{}, as a property of a specific \textit{agent}, or (2) as an abstraction of the EFG representation derived from the underlying \modelAbbrev{}.
In particular, the second option essentially makes all prior EFG research on imperfect recall applicable to our setting as well.

\subsection{Comparison of \modelAbbrev{}s and Classical EFGs}\label{sec:sub:FOSGs-as-EFGs}


We are now ready to make a comparison between \modelAbbrev{}s and EFGs.
We will show that any ``well-behaved'' EFG can be obtained by starting with an extensive-form representation of some \modelAbbrev{} and throwing away all information about what players know outside of their turn (and hence also about public knowledge in the game).
We will also give several examples which show that once lost, the public information is very difficult -- and sometimes even impossible -- to put back.
We start by describing classical EFGs and explaining what we mean when referring to ``well-behaved'' EFGs.

In light of the augmented definition of EFGs, classical EFGs can be defined as follows:

\begin{definition}[Historical definition of EFGs]
A \defword{classical extensive-form game} $E'$ is a tuple $\left< \mc N, \mc A, \mc H, p, \pi_c, \mc I', u \right>$, where all the objects are as in Definition~\ref{def:augmented-EFG}, except that $\mc I'$ is of the form $\mc I' = (\mc I'_1,\dots,\mc I'_N)$, where $\mc I'_i$ only covers $\{ h \in \mc H \mid p(h)=i \}$.
\end{definition}

We thus see that the only \textit{formal} difference between classical and augmented EFGs is the absence of the public partition $\pubSetTree$ and the fact that classical infosets are only defined for the active player.
In particular, any augmented EFG $E$ can be turned into a classical EFG $E' = \texttt{ForgetNonActingI}(E)$ throwing away $E$'s public partition and restricting each $\mc I_i$ to the histories where $i$ acts.
For an intuitive comparison between the two formalisms, we invite the reader to compare Figures~\ref{fig:kuhn_fog} and \ref{fig:kuhn_efg}.

As with augmented EFGs, we can restrict our attention to games with perfect recall.
However, the ``no thick infosets'' condition is both trivial and meaningless (since in a \textit{classical} EFG, infosets are only defined when the player acts, so they cannot be thick unless the player has imperfect recall).
This opens the door to a different pathological property -- non-timeability:

\begin{definition}[Timeability, \cite{timeability}]
For a \textnormal{[classical]} extensive-form game, a deterministic timing is a labelling of the nodes in $\mc H$ with non-negative real numbers such that the label of any node is at least one higher than the label of its parent. 
A deterministic timing is exact if any two nodes in the same information set have the same label.
\end{definition}

\noindent We can then call a game \defword{timeable} if it admits an exact deterministic timing and \defword{1-timeable} if each label is \textit{exactly} one higher than its parent's label.
The difference between timeability and 1-timeability is, however, insignificant:

\begin{lemma}[Equivalence of timeable and 1-timeable EFGs]\label{lem:1-timeability}
Any classical timeable EFG can be made 1-timeable by adding chance nodes with a single \emph{noop} action.
\end{lemma}

In Appendix~\ref{sec:app:padding_size}, we show that this modification will not increase the size of the game more than quadratically.

\begin{proof}[Proof sketch]
This can be shown by first realizing that any exact deterministic timing can, without loss of generality, be integer-valued.
The equivalent 1-timeable game is then obtained by adding a single ``dummy'' chance node for any integer that the timing skipped on the way between some node $ha$ and its parent $h$.
\end{proof}


There are two classes of non-timeable games.
The more benign case are games where one player might take an arbitrarily high (but finite) number of actions between two actions of their opponent (for example, suppose I could keep eating cookies for as long as I keep rolling six on a dice).\footnote{If we allowed for timings that only require the label to be higher (rather than ``at least one higher'') than the parent label, timeability would encompass these games as well.}
The truly obscure case are games whose non-timeability is caused by \textit{cyclic time-dependencies} between nodes. (For example, the game depicted in Figure~\ref{fig:unfair_MP} has histories $g,g' \in I \in \mc I'_1$ and $h,h'\in J \in \mc I'_2$ for which $g$ is a parent of $h$ but $h'$ is a parent of $g'$.)
The authors of \cite{timeability} argue for avoiding non-timeable games because rather than corresponding real-world problems, such games are merely an artefact of the EFG formalism.

\medskip

With this terminology, we are finally ready to describe the relation between \modelAbbrev{} and EFGs.
Denoting by $E' =: \ClassicalEFG{G}$ the classical EFG obtained by turning $E=\EF{G}$ into $E' = \texttt{ForgetNonActingI}(E)$, we have the following:

\begin{restatable}[\modelAbbrev{}s as timeable EFGs]{theorem}{EFGize}\label{thm:FOSGs-as-EFGs}
For any \modelAbbrev{} $G$, the game $E' = \ClassicalEFG{G}$ is a classical 1-timeable perfect-recall EFG.
Moreover, any classical 1-timeable\footnote{By Lemma~\ref{lem:1-timeability}, 1-timeable EFGs have the same expressive power as timeable ones --- this makes the 1-timeability assumption a formality. Moreover, if we were to retract the assumption that players receive observations on \textit{every} transition, we would be able to represent non-timeable games as well. However, the desirability of doing so is debatable.} perfect-recall\footnote{Regarding imperfect recall in \modelAbbrev{}s, see the remark below Theorem~\ref{thm:FOSG-to-aug-EFG}.}
EFG can be obtained this way.
\end{restatable}

In the remainder of this section, we look at some of the downsides of classical EFGs.
    \subsection{Unsuitability of Classical EFGs for Decomposition}\label{sec:EFG-problems}

\begin{figure*}[tb]
        \centering
        \scalebox{0.75}{
            \rotatebox{90}{
                \setlength{\nodesize}{2em}
                \begin{tikzpicture}
                \pokerHistory
                \pokerAA
                \end{tikzpicture}
                \begin{tikzpicture}
                \pokerHistory
                \pokerB
                \end{tikzpicture}
                \begin{tikzpicture}
                \pokerHistory
                \pokerC
                \end{tikzpicture}
                \begin{tikzpicture}
                \pokerHistory
                \pokerDD
                \end{tikzpicture}
            }
        }
    \caption[\cite{ObsBasedEFGmodel}]{Bottom to top:
        (a): A subset of the public partition \emph{as we intuitively imagine it} in the game of limit Texas hold 'em, corresponding to a history where both players only placed the default bets.
        (b) and (c): Classical (solid) and augmented (dashed) information sets of player 1 and player 2, as defined by \cite{CFR-D}.
        (d): The public states \emph{actually} corresponding to these information partitions.
        There are only five public states along this history (rather than 12 as one might intuitively expect), and they do not even correspond to the rounds of the game.
    }
    \label{fig:poker_I_aug}
\end{figure*}

The classical EFG formalism has several important problems that stem from one crucial difference between ``classical-EFG-ization'' (Theorem~\ref{thm:FOSGs-as-EFGs}) and ``augmented-EFG-ization'' (Theorem~\ref{thm:FOSG-to-aug-EFG}):
While Theorem~\ref{thm:FOSG-to-aug-EFG} relies on the 1-to-1 correspondence between extensive-form representations and augmented EFGs, the mapping $\texttt{ForgetNonActiveI}$ between augmented and classical EFGs is very much \textit{not} 1-to-1.
There are various ways to see this.
One indication is that the construction we used to prove the ``moreover'' part of Theorem~\ref{thm:FOSGs-as-EFGs} is far from canonical, so there might be various vastly different augmented EFGs or \modelAbbrev{}s which nevertheless correspond to the same classical EFG.
When attempting to augment a classical EFG, we start with the information players have when they are active and attempt to reverse-engineer \textit{when} was this information obtained.
However, the following examples suggest that there is no canonical solution to this issue.
Indeed, on the one end, there are games where players receive most of their information from the actions of \textit{others} (e.g., Hanabi). Then there are games where each player's actions reveal information to everybody (e.g., blind chess or battleships), but also games where information is primarily gained through one's own actions (e.g., turn-based map-exploration games such as Heroes of Might and Magic).
Worse yet, Figure~\ref{fig:no_finest_partition} shows two augmented EFGs which share the history tree and correspond to the same classical EFG despite having different information partitions and public sets.

\begin{figure}[tb]
        \centering
        \scalebox{0.5}{
            \begin{tikzpicture}
            \unfairMP
            \unfairMPtext
            \end{tikzpicture}
        }
        \scalebox{0.75}{
            \setlength{\nodesize}{1.5em}
            \begin{tikzpicture}[scale=1]
            \noFinestPartitionExample
            \noFinestPartitionExampleA
            \end{tikzpicture}
            \begin{tikzpicture}[scale=1]
            \noFinestPartitionExample
            \noFinestPartitionExampleB
            \end{tikzpicture}
        }
    \caption{
        Left:
            A non-timeable classical EFG (from \cite{ObsBasedEFGmodel}). The circle denotes a chance node, the triangles denote different players, rectangles denote classical infosets.
        Right:
            Two different augmented EFGs (resp. their parts) that correspond to the same classical EFG \cite{ObsBasedEFGmodel}. Classical information sets are solid, the augmented information sets (i.e. the information sets of the non-active player) are dashed.
    }
    \label{fig:no_finest_partition}\label{fig:unfair_MP}
\end{figure}

This proves that classical EFGs cannot possibly offer any \textit{canonical} notion of public information. 
However, even if we were to leave canonicity aside, Figure~\ref{fig:poker_I_aug} shows that the best existing attempt \cite{CFR-D} to reconstruct public information from classical EFGs leads to extremely unintuitive results in poker.\footnote{As the author of this paragraph can attest, this gets particularly confusing when one is not aware of the fact that their algorithms and informal thoughts use the ``intuitive'' augmentation of the classical EFG while the corresponding formal analysis uses the unintuitive one.} For a more formal analysis of similar issues with classical EFGs, see our previous technical report \cite{ObsBasedEFGmodel}.
\section{Transfer of Results Between Game Theory and Multiagent RL}\label{sec:translating_results} 
    As noted in the introduction, problems studied in the EFG and POSG models are closely related.
We give several examples of POSG results which might inform future work in EFGs.
In the opposite direction, we use \modelAbbrev{}s to translate three influential EFG results into the \modelAbbrev{} language that might be more accessible to researchers with RL background.
We start with counterfactual regret minimization, the now-standard algorithm which lies at the heart of recent successes such as \cite{DeepStack,Pluribus}.
We then talk about decomposition in \modelAbbrev{}s and illustrate it using CFR-D, the decomposition variant of CFR.
The third method we discuss is sequence form, a tool with which we can apply linear programming to \modelAbbrev{}s.

\subsection{Examples of MARL results related to EFGs}

There are multiple results in the Dec-POMDP which are highly relevant to EFGs.
For example, \cite{oliehoek2009lossless} studies lossless clustering of histories in Dec-POMDPs, which very closely corresponds to the notion of well-formed imperfect-recall games \cite{lanctot2012,CRSWFG}.
Similarly, the notion of sufficient plan-time statistics studied in \cite{oliehoek2013sufficient} is very closely related to the inputs to the value function used in DeepStack \cite{DeepStack}, and the paper suggests the structure of the value function for games where the hidden information becomes public after a certain number of moves.
\cite{dibangoye2016optimally} proposes alternative sufficient statistics (and their compact representation) which rely on the underlying world states.
Since this idea does not seem to be present in any of the existing EFG literature, this work offers inspiration for future research.

\subsection{Counterfactual Regret Minimization}\label{sec:CFR}
Counterfactual Regret Minimization (CFR)~\citep{07nips-cfr} is an algorithm for approximately solving two-player, zero-sum games by finding an approximation of a Nash equilibrium. It fits into the general category of a policy improvement algorithm and can be compactly described as self-play with independent instances of the regret-matching algorithm~\citep{hart2000simple} operating over the actions at each information state for each player, using a particular notion of expected reward for the actions, the counterfactual value.
We now describe how CFR looks when applied to (the extensive-form representation of) \modelAbbrev{}s.
We then show that this results in the same strategy we would obtain by running CFR on the corresponding EFGs, which guarantees that all the existing results from the CFR literature translate to \modelAbbrev{}s.


In \modelAbbrev{}s, we define the \defword{value} $v^\pi_i(h)$ \defword{of a history} $h=\initState \ldots w \in \mc H$ as the expected future reward --- i.e., the cumulative reward $\mc R_i(\tau)$ along the trajectory $\tau = w a \ldots $ that starts at the endpoint of $h$ and ends in some terminal world state:
\begin{equation*}
    v^\pi_i(h) := \mathop{\mathbb{E}} \left[ \mc R_i(\tau) \mid \tau \textnormal{ generated from $\pi$ and $\mc T$, current history is } h \right].
\end{equation*}
The \defword{value} $q^{\pi}_i(h,a_i)$ \defword{of an action} $a\in \mc A_i(h)$ is defined analogously, except that $i$ is additionally assumed to take $a_i$ at $h$.
These definitions can be extended to \defword{values of information states} (or sets) by defining
\begin{equation}
    v^\pi_i(s_i) := \sum_{s_i(h)=s_i} P^\pi(h|s_i) v^\pi_i(h), \ \ \ \ 
    q^\pi_i(s_i,a) := \sum_{s_i(h)=s_i} P^\pi(h|s_i) q^\pi_i(h,a_i) ,
\end{equation}
where $P^\pi(h|s_i)$ denotes the probability that $h$ is the current history when $s_i$ is $i$'s current information state.\footnote{To avoid issues with division by zero, we can rewrite $P^\pi(h|s_i)$ in terms of counterfactual reach probabilities: $P^\pi(h|s_i) = P^\pi(h) / P^\pi(s_i) = P^\pi_{-i}(h) / P^\pi_{-i}(s_i)$. For a more detailed explanation (including the treatment of the case where $P^\pi_{-i}(s_i) = 0$), see \cite{seitzDLS}.}

Finally, \defword{counterfactual values} $v^\pi_{i,\textnormal{cf}}(s_i) = P^\pi_{-i}(s_i) v^\pi_i(s_i)$ and $q^\pi_{i,\textnormal{cf}}(s_i,a_i) = P^\pi_{-i}(s_i) q^\pi_i(s_i,a_i)$ are defined by scaling values proportionally to the probability that $s_i$ is reached in the counterfactual scenario where $i$ plays to reach $s_i$.
(Formally, the counterfactual reach probability of $s_i$ is $P_{-i}^\pi(s_i) := \sum_{s_i(h)=s_i} P_{-i}^\pi(h)$, where $P^{\pi}_{-i}(h) := P_{c}(h)\prod\nolimits_{j \in \mc N \setminus \{i\}} P^{\pi}_j(h)$ is defined in terms of the contributions $P_{c}(h) := \prod_{h'aw \sqsubseteq h}\mc{T}(h',a,w)$ of chance and $P^{\pi}_{j}(h) := \prod_{h'aw \sqsubseteq h}\pi_j(s_j(h'),a_j)$ of player $j$ to the probability of reaching $h$.)


Given a strategy sequence $\pi^0, \dots, \pi^{t-1}$, the counterfactual regret at time $t$ for an action $a$ at state $s$ is
\begin{align}\label{eq:regret}
R^t_i(s_i,a_i) =
\sum\nolimits_{k=0}^{t-1} \left( q^{\pi^k}_{i,\textnormal{cf}}(s_i,a_i) - v^{\pi^k}_{i,\textnormal{cf}}(s_i) \right)
\end{align}
Denoting $(x)^+:=\max(x,0)$, the \defword{regret-matching policy} $\pi^{t}(s_i,a_i)$ at time $t$ is defined as
\begin{align*}
\pi^{t}(s_i,a_i) := (R^t_i(s_i,a_i))^+ / \sum\nolimits_{a' \in \mc{A}(s)} (R^t_i(s,a'))^+ ,
\end{align*}
resp. as $\pi^{t}(s_i,a_i) := \edit{1 / |\mc{A}(s_i)|}$ when the denominator is $0$.

In the EFG setting, CFR is defined analogously, except for using utilities $u_i(z)$ to define $v_i^\pi(h)$ (i.e., \textit{total} cumulative rewards rather than \textit{future} rewards).
The following result shows that if a \modelAbbrev{} has observable rewards, the difference between ``EFG-values'' and ``\modelAbbrev{}-values'' cancels out, which yields to following result:

\begin{restatable}[CFR in \modelAbbrev{}s]{theorem}{CFR}\label{thm:CFR}
Suppose that a \modelAbbrev{} G has observable rewards, i.e., that $\mc R_i(h) = \mc R_i(h')$ whenever $s_i(h)=s_i(h')$.
Then running CFR on $G$ yields the same strategy as running CFR on $E' = \ClassicalEFG{G}$.
\end{restatable}

In particular, we get an analogue of Zinkevich et al.'s result \cite{07nips-cfr} according to which the profile of average policies generated by CFR forms an $\epsilon$-Nash equilibrium for some suitable $\epsilon = \mc{O}(\max_i |\mc{S}_i| \sqrt{|\mc{A}|_i}/\sqrt{t})$ whenever a game is two-player zero-sum and has perfect recall.

    \subsection{Decomposition and CFR-D}\label{sec:cfr-d}

Decomposition in imperfect information games is conceptually more complicated than its perfect-information variant \cite{CFR-D,DeepStack,moravcik2016refining,brown2018depth,accelerated_BR,seitzDLS,MCCR}.
Rather than attempting to give a general theory of decomposition in \modelAbbrev{}s, we thus describe a specific example:
CFR-D, a -Decomposition variant of the standard CFR, together with basic definitions needed to use it in FOSGs.


In \modelAbbrev{}s, decomposition revolves around subtrees of the public state-tree $\pubTree$ --- i.e., with sets of the form $\textnormal{Desc}(\pub) := \{ \pub' \in \pubTree \mid \pub' \sqsupset \pub \}$ for some $\pub\in \pubTree$.
These subtrees also implicitly generate sub-trees (or rather, forests) within the space $\mc I_i$ of information states and the space $\mc H$ of histories.\footnote{Furthermore, both the public state tree and the information state forest can be identified with the corresponding tree (resp. forest) within the public \textit{set} tree and the information \textit{set} tree.}
To gain the basic intuition for the equivalence of these views, we refer the reader to Figure~\ref{fig:kuhn_fog}.
The formal relationship is then captured by Proposition~\ref{prop:subgame}.


A basic form of actual decomposition is thus to pick some \defword{trunk} $\mc T$ of $\pubTree$ (a sub-tree $\mc T \subset \pubTree$ that has the same root as $\pubTree$).
Denoting its \defword{leaves} as $\mc L := \{ \pub \in \pubTree \mid \pub \textnormal{ lies just below } \mc T \}$, we decompose $\pubTree$ into $\mc T$ and $\textnormal{Desc}(\pub)$, $\pub \in \mc L$.

However, to turn the sets $\textnormal{Desc}(\pub)$ into standalone games that can be solved, we need to add extra information.
For the purpose of CFR-D, it suffices to assume that the strategies used in the trunk are public knowledge and note down the corresponding distributions over information states in the trunk's leaves (for further details, see \cite{seitzDLS}):

\begin{definition}[Range]
For $\pub\in\pubTree$, a tuple $r = (r_1,\dots,r_N) \in \Delta \mc S_1(\pub) \times \dots \times \Delta \edit{\mc S_N}(\pub)$ is called a (normalized) \defword{range} at $\pub$.
\end{definition}

\noindent 
\edit{For example, suppose that the strategy of Player 1 in Kuhn poker is to bet with probability $1/16$, $3/16$, resp. $1$ when holding $J$, $Q$, resp. $K$,
and $\pub$ is the public state where Player 1 just bet.
Then the non-normalized range of Player 1 is\footnote{\edit{It is debatable whether range should include the contribution of chance or not.
This can make a difference in general games where -- unlike in our example -- the chance distribution is non-uniform.
However, even in general, the two representations are equivalent since the chance distribution can be derived from the game's rules and factored out \cite{seitzDLS}.}}
$ r'_1 = ( \frac{1}{16}, \frac{3}{16}, 1 ) $ and normalizing it (i.e., dividing by $\frac{1}{16} + \frac{3}{16} + \frac{16}{16}$) yields $r_1 = (5\%, 15\%, 80\%)$.
Ranges are closely connected to the notion of belief from the POSG literature:
Suppose that player two has card $K$ and knows $r_1$.
Updating on the fact that player one cannot have $K$ leads to the belief that player one has $J$ with $\frac{5}{5+15} = 25\%$ probability and $Q$ with $\frac{15}{5+15} = 75\%$ probability.
More generally, we can informally say that a belief of Player $p$ allows us to compute the ranges $r_i$ of all other players, and vice versa.}
Formally, the concept of range enables the following useful terminology (which also applies to non-normalized ranges):

\begin{definition}[Public belief state \cite{ReBeL}]
For $\pub \in \pubTree$ and range $r$ at $\pub$, the pair $\beta = (\pub,r)$ is called a \defword{public belief state} (PBS).
\end{definition}

For a PBS $\beta$ in a \modelAbbrev{} $G$, a \defword{subgame $G(\beta)$ rooted at a public belief state $\beta$} is defined as a \modelAbbrev{} that is identical to $G$, except that  it ``starts at $\pub$, with the initial range $r$''.\footnote{Strictly speaking, this type of subgame should be called an \textit{value-solving} subgame, because its purpose is to compute the value of the subgame, not to recover an optimal strategy for that part of the game tree. However, for the purpose of CFR-D, this type of subgame is sufficient \cite{Neil_thesis}.}
In Appendix~\ref{sec:app:decomposition}, we formally construct $G(\beta)$ by adding a new initial world-state to $G$.
This initial state emulates the trunk portion of $G$, sampling a single outcome and producing corresponding rewards and observations.
On top of this, it also sends the range $r$ to all players as a public observation.\footnote{This formally allows game-solving algorithms to utilize this information. Otherwise, the algorithms would have to treat all actions taken in the trunk as produced by chance.}

\defword{CFR-D} is an algorithm designed to serve as a slower but less memory-hungry alternative to the classical CFR \cite{CFR-D} (Algorithm~\ref{alg:cfr-d}).
It achieves this by performing the forward-pass of CFR in a trunk $\mc T$ and then solving each leaf-subgame $G(\pub,r)$, $\pub \in \mc L$, with respect to the range corresponding to the current strategy.
Afterwards, the algorithm recovers the infoset values at the roots of these subgames and uses them for the CFR backwards-pass in the trunk.
This process repeats for a specified number of iterations, after which the algorithm outputs the average of the trunk strategies across all iterations.

\begin{algorithm}[t]
\SetKwInOut{Input}{Input}
\SetKwInOut{Output}{Output}
\LinesNumbered
\Input{\modelAbbrev{} $G$, Trunk $\mc T \subset \mc \pubTree$, $T\in \N$}
\Output{A trunk strategy $\bar \pi^T$}
\BlankLine

$R^0$ $\leftarrow$ 0, 
$\pi^0|_{\mc T}$ $\leftarrow$ uniformly random strategy\;
\For{$t=0,\dots,T-1$}{
	$P$ $\leftarrow$ compute reach probabilities in $\mc T$ and $\mc L$ under $\pi^t|_{\mc T}$\;
	\For{$\pub \in \mc L$}{
		$r$ $\leftarrow$ range at $\pub$ corresponding to $P$\;
		$\pi^t|_{\textnormal{Desc}(\pub)}$ $\leftarrow$ solve $G(\pub,r)$\;
		$\vec v_{\pub}$ $\leftarrow$ the vector of cf. values $v^{\pi^t}_{i,\textnormal{cf}}(s_i)$ for $s_i$ at $\pub$\;
	}
	$R^{t+1}$ $\leftarrow$ update regrets $R^t$ in $\mc T$ using $P^t$ and $\vec v_{\pub}$, $\pub \in \mc L$\;
	$\pi^{t+1}|_{\mc T}$ $\leftarrow$ apply Regret Matching to $R^{t+1}$\;
}
\textbf{return} $\bar \pi^T = \frac{1}{T}\sum_{t=1}^T \pi^t|_{\mc T}$\;
\caption{CFR-D, the decomposition variant of CFR}\label{alg:cfr-d}
\end{algorithm}
    \subsection{\modelAbbrev{} Sequence Form}\label{sec:app:sequence_form}

Sequence-form representation is a prominent method in EFGs, as it allows for effective solving of two-player zero-sum EFGs by linear programming \cite{MAS,nisan2007algorithmic}.
In this section, we describe a sequence-form representation of a \modelAbbrev{}.
We keep the structure and notation close to the sequence form representation of EFGs as presented in \cite{nisan2007algorithmic}.


We use $\mathcal{B}_i := \{ sa \mid s \in \mathcal{S}_i, a \in \mathcal{A}_i(s) \}$ to denote the set of all action-observation sequences of a player $i$ that end in an action and set $b_i(ha) := s_i(h)a$.
For a fixed policy $\pi$, the sequence form \defword{realization probability} of a sequence $sa \in \mathcal{B}_i$ is $\prod_{s'a' \sqsubseteq sa} \pi_i(s', a')$.
In a two-player \modelAbbrev{}, we use $x$ and $y$ to denote the vectors of these quantities for all such sequences of player $1$ and $2$ respectively.
The \defword{sequence-form payoff matrix} $A \in \mathbb{R}^{\mathcal{B}_1 \times \mathcal{B}_2}$ of the first player consists of entries $A_{\sigma \tau}$ for $\sigma \in \mathcal{B}_1, \tau \in \mathcal{B}_2$ where
\begin{align}
    A_{\sigma \tau} = \sum_{z = haw \in \mc Z \, : \, b_1(ha) = \sigma, b_2(ha) = \tau } P_{\mathcal{T}}(z) u_1(z) .
\end{align}
Using this notation, the expected utility under $\pi$ can be rewritten in terms of realization plans as
\begin{align}
 \mathbb{E}_z \left[ u_i(z) \mid \pi \right] = x^\top A y .
\end{align}

Analogously to \cite[Proposition~3.10]{nisan2007algorithmic}, one can show that a vector $x=(x_{sa})_{sa\in \mc B_i}$ is a realization plan corresponding to some policy $\pi$ if and only if it is non-negative and satisfies the following two properties:
\begin{align}
    x_{\emptyset} &= 1 \label{eq:root=1}\\
    \sum\nolimits_{a' \in \mathcal{A}(sao)} x_{saoa'} &= x_{sa} .\label{eq:realiz_plan}
\end{align}
Reorganizing \eqref{eq:root=1} and \eqref{eq:realiz_plan}, we can be compactly express these equations using sparse $\pm 1$-valued matrices $E, F$ and vectors $e, f$ \cite[Section 3.10]{nisan2007algorithmic}:
\begin{align}
    Ex = e, \, x & \geq \boldsymbol{0} \\
    Fy = f, \, y & \geq \boldsymbol{0} .
\end{align}
Similarly to \cite[Theorem~3.13]{nisan2007algorithmic}, an approximate NE of a two-player perfect-information zero-sum \modelAbbrev{} can by computed by the LP
\begin{equation*}
\underset{u,y}{\textnormal{minimize}} \ e^T u, \ \textnormal{subj.\,to } Fy=f, \ E^T u - Ay \geq \boldsymbol{0}, \ y\geq \boldsymbol{0} .
\end{equation*}
\section{\edit{Discussion}}\label{sec:disc}
    \edit{Finally, we summarize our contribution (Section~\ref{sec:sub:conclusion}) and contrast it with the most-relevant prior work (Section~\ref{sec:related_work}).}
    \subsection{\edit{Related Work}}\label{sec:related_work}
\edit{First, let us distinguish the proposed factorization -- that of \textit{observations} into private and public -- from other types of factorization in the literature.
First, Factored Dec-POMDPs \cite{factoredDecPOMDP1, factoredDecPOMDP2} consider situations where
the \textit{state} is factored into variables $s = (x_1, \dots, x_k)$ and
the reward function can be written as a sum $\mc R = \sum_{e=1}^k \mc R^e$, where each ``local'' reward $\mc R^e$ depends only on $x_e$.
(A motivating example is a situation where firefighters want to minimize the damage from simultaneous fires in different locations.)
Second, \cite{factoredVF2} attempt to decompose the \textit{value function} into a sum of functions which only depend on the actions of a \textit{subset of players}.
For example, in the above setup, an ambulance and firefighters (players) could be at the same location ($x_e$) while having largely independently effects (on the value function).
Third, the factorization of states into variables is also leveraged in Factored MDPs, where it is used to represent the MDP's dynamics more compactly \cite{factoredMDP}.
Overall, all three approaches seem to address different challenges than the \modelAbbrev{} model while being compatible with it where applicable.
Finally, \cite{nayyar2013decentralized} study decentralized stochastic control with a coordinator who knows all common information.
Of all the factorizations we know of, this one is the closest in spirit to \modelAbbrev{}s.
However, it uses language very different from ours and only considers fully cooperative settings.
}

\edit{Since a significant contribution of this text is conceptual -- the description of \modelAbbrev{}s and their extensive-form representation -- let us review the origin of some of the model's ingredients.}
\edit{The idea of considering \textit{information sets for non-acting players} appears first in \cite{CFR-D}, where ``augmented'' infosets are derived from the structure of the underlying EFG.
Infosets that partition the whole game tree are also implicitly used in \cite{brown2018depth}, where they are assumed to be given as part of the description of the EFG.
Finally, \cite{ObsBasedEFGmodel} points out the impossibility of deriving the ``correct'' infosets from the underlying EFG and suggests explicitly deriving infosets from observations.}
\edit{Another related concept are \textit{public states}, inspired by public cards in poker.
The notion was first defined for general EFGs in \cite{accelerated_BR},
and \cite{MCCR} later gave a definition in terms of closure under everywhere-defined infosets.
\cite{MCCR} further pointed out that the definition is ambiguous and defined \textit{common-knowledge states} as \textit{the finest possible public states}.}
\edit{The idea of decomposing EFGs into subgames dates at least to \cite{billings2003approximating}.
As far as we know, \cite{CFR-D} is the first to define subgames formally, using (what is now called) common-knowledge states as subgame roots.
The latest addition in this vein was the introduction of \textit{public belief states} by \cite{ReBeL}.
A part of our contribution is consolidating these concepts into a single model.
(But we also give other results, such as establishing the connection between POSGs/\modelAbbrev{}s and EFGs.)}

\edit{To explain the relationship between our contribution and the literature around CFR \cite{CFR},
note that both the CFR literature and the above-outlined conceptual progress are tied to algorithmic progress in combining search and reinforcement learning.
This line of research produced a number of results, including 
CFR-D \cite{CFR-D} (the decomposition-variant of CFR), DeepStack \cite{DeepStack} (learns to play poker on a superhuman level without human input), and ReBeL \cite{ReBeL} (which also learns games without human input, but isn't limited to poker).
Throughout all this research, the conceptual progress was both motivated by the desire to develop better algorithms and vital for doing so.
This is particularly true for the present paper:
The \modelAbbrev{}s model was developed as a response to the realization that classical EFGs are insufficient for extending DeepStack beyond poker \cite{ObsBasedEFGmodel}.
And soon after its appearance\footnote{\edit{This circular citation is possible since the first draft of this text appeared in 2019.}}, the model was adopted as the underlying formalism for ReBeL.}

\edit{Finally, \modelAbbrev{}s can be viewed as a continuation of the trend of combining algorithmic game theory -- and CFR variants in particular -- with multiagent RL.
We would particularly like to highlight the paper \cite{srinivasan2018actor}, which first expressed a number of the CFR-related concepts in the notation traditionally used in MARL.
Since the number of recent results on the intersection of algorithmic game theory and MARL is large, reviewing them in more detail would be beyond the scope of this text.}
    \subsection{Conclusion}\label{sec:sub:conclusion}

The two main takeaways of this work are the following:
First, \textit{to get the most value out of game-theoretical models, we should no longer discard information about whether observations in an environment are observed jointly or privately.}
We propose that this can be done by using \modelName{}s (\modelAbbrev{}s), a minor extension to the existing partially-observable stochastic game model.
Second, EFGs and POSGs/\modelAbbrev{}s should not be viewed as two unrelated models.
Instead, \textit{EFGs are derived objects that can be obtained by ``unrolling'' some underlying \modelAbbrev{}.}
This both highlights a new application for standalone EFG research and suggests that explicitly reasoning about the underlying \modelAbbrev{} can have significant benefits.
(Indeed, \modelAbbrev{}s have the potential to enable new techniques and insights, make results accessible to a broader audience, and align our formal language with recent domain-implementations which often already resemble \modelAbbrev{}s.)

More specifically, we have made the following contributions:
We show an equivalence between POSGs and \modelAbbrev{}s.
Any POSG result directly applies to \modelAbbrev{}s as well, since it suffices to merely forget the factorization of observations (Proposition~\ref{prop:POSG-ization}).

We provide a mapping between EFGs and \modelAbbrev{}s.
Every \modelAbbrev{} has a canonical\footnote{Up to the order of players in the serialization step, see Section~\ref{sec:sub:efr}.} extensive-form representation, and every well-behaved (perfect recall and timeable) EFG corresponds to some \modelAbbrev{} (Theorems~\ref{thm:FOSG-to-aug-EFG} and \ref{thm:FOSGs-as-EFGs}).
Moreover, this relationship between \modelAbbrev{}s and EFGs suggests that our Definition~\ref{def:augmented-EFG} is a natural way of formalizing EFGs.
As a combination of several recent extensions of the EFG model \cite{CFR-D,accelerated_BR,brown2018depth,seitzDLS}, our formalization removes the need for implementing modern search algorithms in a domain-specific manner.
In particular, it preempts various problems with decomposition that are difficult or even impossible to solve in the historical formalization of EFGs (Section~\ref{sec:EFG-problems}).
%
%

Finally, we demonstrate that translating EFG results to \modelAbbrev{}s is straightforward --- decomposition in \modelAbbrev{}s has intuitive properties (Section~\ref{sec:cfr-d}) and two key EFG techniques -- counterfactual regret minimization and sequence form -- are easy to formulate in this framework (Sections~\ref{sec:CFR} and \ref{sec:app:sequence_form}).

\subsubsection*{Acknowledgements}
  We are grateful to Tomáš Gavenčiak, Karel Horák, and Marc Lanctot for discussions related to an earlier version of this paper.
This work was supported by Czech Science Foundation grant no. 18-27483Y and OP VVV MEYS funded project CZ.02.1.01/0.0/0.0/16 019/0000765 Research Center for Informatics.

\bibliographystyle{elsarticle-harv}
\bibliography{references}

\begin{thebibliography}{52}
\expandafter\ifx\csname natexlab\endcsname\relax\def\natexlab#1{#1}\fi
\providecommand{\url}[1]{\texttt{#1}}
\providecommand{\href}[2]{#2}
\providecommand{\path}[1]{#1}
\providecommand{\DOIprefix}{doi:}
\providecommand{\ArXivprefix}{arXiv:}
\providecommand{\URLprefix}{URL: }
\providecommand{\Pubmedprefix}{pmid:}
\providecommand{\doi}[1]{\href{http://dx.doi.org/#1}{\path{#1}}}
\providecommand{\Pubmed}[1]{\href{pmid:#1}{\path{#1}}}
\providecommand{\bibinfo}[2]{#2}
\ifx\xfnm\relax \def\xfnm[#1]{\unskip,\space#1}\fi
\bibitem[{Badia et~al.(2020)Badia, Piot, Kapturowski, Sprechmann, Vitvitskyi,
  Guo and Blundell}]{agent57}
\bibinfo{author}{Badia, A.P.}, \bibinfo{author}{Piot, B.},
  \bibinfo{author}{Kapturowski, S.}, \bibinfo{author}{Sprechmann, P.},
  \bibinfo{author}{Vitvitskyi, A.}, \bibinfo{author}{Guo, D.},
  \bibinfo{author}{Blundell, C.}, \bibinfo{year}{2020}.
\newblock \bibinfo{title}{Agent57: Outperforming the atari human benchmark}.
\newblock \bibinfo{journal}{arXiv preprint arXiv:2003.13350} .
\bibitem[{Billings et~al.(2003)Billings, Burch, Davidson, Holte, Schaeffer,
  Schauenberg and Szafron}]{billings2003approximating}
\bibinfo{author}{Billings, D.}, \bibinfo{author}{Burch, N.},
  \bibinfo{author}{Davidson, A.}, \bibinfo{author}{Holte, R.},
  \bibinfo{author}{Schaeffer, J.}, \bibinfo{author}{Schauenberg, T.},
  \bibinfo{author}{Szafron, D.}, \bibinfo{year}{2003}.
\newblock \bibinfo{title}{Approximating game-theoretic optimal strategies for
  full-scale poker}, in: \bibinfo{booktitle}{IJCAI}, pp.
  \bibinfo{pages}{661--668}.
\bibitem[{Boutilier et~al.(2000)Boutilier, Dearden and
  Goldszmidt}]{factoredMDP}
\bibinfo{author}{Boutilier, C.}, \bibinfo{author}{Dearden, R.},
  \bibinfo{author}{Goldszmidt, M.}, \bibinfo{year}{2000}.
\newblock \bibinfo{title}{Stochastic dynamic programming with factored
  representations}.
\newblock \bibinfo{journal}{Artificial intelligence} \bibinfo{volume}{121},
  \bibinfo{pages}{49--107}.
\bibitem[{Brown et~al.(2020)Brown, Bakhtin, Lerer and Gong}]{ReBeL}
\bibinfo{author}{Brown, N.}, \bibinfo{author}{Bakhtin, A.},
  \bibinfo{author}{Lerer, A.}, \bibinfo{author}{Gong, Q.},
  \bibinfo{year}{2020}.
\newblock \bibinfo{title}{Combining deep reinforcement learning and search for
  imperfect-information games}.
\newblock \bibinfo{journal}{arXiv preprint arXiv:2007.13544} .
\bibitem[{Brown and Sandholm(2017a)}]{brown2017safe}
\bibinfo{author}{Brown, N.}, \bibinfo{author}{Sandholm, T.},
  \bibinfo{year}{2017}a.
\newblock \bibinfo{title}{Safe and nested subgame solving for
  imperfect-information games}, in: \bibinfo{booktitle}{Advances in Neural
  Information Processing Systems}, pp. \bibinfo{pages}{689--699}.
\bibitem[{Brown and Sandholm(2017b)}]{Libratus}
\bibinfo{author}{Brown, N.}, \bibinfo{author}{Sandholm, T.},
  \bibinfo{year}{2017}b.
\newblock \bibinfo{title}{{Superhuman {AI} for heads-up no-limit poker:
  {L}ibratus beats top professionals}}.
\newblock \bibinfo{journal}{Science} , \bibinfo{pages}{eaao1733}.
\bibitem[{Brown and Sandholm(2019)}]{Pluribus}
\bibinfo{author}{Brown, N.}, \bibinfo{author}{Sandholm, T.},
  \bibinfo{year}{2019}.
\newblock \bibinfo{title}{Superhuman {AI} for multiplayer poker}.
\newblock \bibinfo{journal}{Science} \bibinfo{volume}{365},
  \bibinfo{pages}{885--890}.
\bibitem[{Brown et~al.(2018)Brown, Sandholm and Amos}]{brown2018depth}
\bibinfo{author}{Brown, N.}, \bibinfo{author}{Sandholm, T.},
  \bibinfo{author}{Amos, B.}, \bibinfo{year}{2018}.
\newblock \bibinfo{title}{Depth-limited solving for imperfect-information
  games}, in: \bibinfo{booktitle}{Proceedings of the 32nd International
  Conference on Neural Information Processing Systems}, pp.
  \bibinfo{pages}{7674--7685}.
\bibitem[{Burch(2017)}]{Neil_thesis}
\bibinfo{author}{Burch, N.}, \bibinfo{year}{2017}.
\newblock \bibinfo{title}{Time and space: Why imperfect information games are
  hard}.
\newblock Ph.D. thesis. University of Alberta.
\bibitem[{Burch et~al.(2014)Burch, Johanson and Bowling}]{CFR-D}
\bibinfo{author}{Burch, N.}, \bibinfo{author}{Johanson, M.},
  \bibinfo{author}{Bowling, M.}, \bibinfo{year}{2014}.
\newblock \bibinfo{title}{Solving imperfect information games using
  decomposition.}, in: \bibinfo{booktitle}{AAAI}, pp.
  \bibinfo{pages}{602--608}.
\bibitem[{{\v{C}}erm{\'a}k et~al.(2017){\v{C}}erm{\'a}k, Bo{\v{s}}ansk{\'{y}}
  and Lis{\'{y}}}]{cermak2017ijcai}
\bibinfo{author}{{\v{C}}erm{\'a}k, J.}, \bibinfo{author}{Bo{\v{s}}ansk{\'{y}},
  B.}, \bibinfo{author}{Lis{\'{y}}, V.}, \bibinfo{year}{2017}.
\newblock \bibinfo{title}{An algorithm for constructing and solving imperfect
  recall abstractions of large extensive-form games}, in:
  \bibinfo{booktitle}{Proceedings of the 26th International Joint Conference on
  Artificial Intelligence}, \bibinfo{organization}{AAAI Press}. pp.
  \bibinfo{pages}{936--942}.
\bibitem[{Davis et~al.(2019)Davis, Schmid and Bowling}]{davis2019low}
\bibinfo{author}{Davis, T.}, \bibinfo{author}{Schmid, M.},
  \bibinfo{author}{Bowling, M.}, \bibinfo{year}{2019}.
\newblock \bibinfo{title}{Low-variance and zero-variance baselines for
  extensive-form games}.
\newblock \bibinfo{journal}{arXiv preprint arXiv:1907.09633} .
\bibitem[{Dibangoye et~al.(2016)Dibangoye, Amato, Buffet and
  Charpillet}]{dibangoye2016optimally}
\bibinfo{author}{Dibangoye, J.S.}, \bibinfo{author}{Amato, C.},
  \bibinfo{author}{Buffet, O.}, \bibinfo{author}{Charpillet, F.},
  \bibinfo{year}{2016}.
\newblock \bibinfo{title}{Optimally solving {Dec-POMDP}s as continuous-state
  {MDP}s}.
\newblock \bibinfo{journal}{Journal of Artificial Intelligence Research}
  \bibinfo{volume}{55}, \bibinfo{pages}{443--497}.
\bibitem[{Emery-Montemerlo et~al.(2004)Emery-Montemerlo, Gordon, Schneider and
  Thrun}]{POSGsCommPayoff2004emery}
\bibinfo{author}{Emery-Montemerlo, R.}, \bibinfo{author}{Gordon, G.},
  \bibinfo{author}{Schneider, J.}, \bibinfo{author}{Thrun, S.},
  \bibinfo{year}{2004}.
\newblock \bibinfo{title}{Approximate solutions for partially observable
  stochastic games with common payoffs}, in: \bibinfo{booktitle}{Proceedings of
  the Third International Joint Conference on Autonomous Agents and Multiagent
  Systems, 2004. AAMAS 2004.}, \bibinfo{organization}{IEEE}. pp.
  \bibinfo{pages}{136--143}.
\bibitem[{Everitt and Hutter(2016)}]{everitt2016avoiding}
\bibinfo{author}{Everitt, T.}, \bibinfo{author}{Hutter, M.},
  \bibinfo{year}{2016}.
\newblock \bibinfo{title}{Avoiding wireheading with value reinforcement
  learning}, in: \bibinfo{booktitle}{International Conference on Artificial
  General Intelligence}, \bibinfo{organization}{Springer}. pp.
  \bibinfo{pages}{12--22}.
\bibitem[{Fagin et~al.(2003)Fagin, Moses, Halpern and
  Vardi}]{fagin2003reasoning}
\bibinfo{author}{Fagin, R.}, \bibinfo{author}{Moses, Y.},
  \bibinfo{author}{Halpern, J.Y.}, \bibinfo{author}{Vardi, M.Y.},
  \bibinfo{year}{2003}.
\newblock \bibinfo{title}{Reasoning about knowledge}.
\newblock \bibinfo{publisher}{MIT press}.
\bibitem[{Foerster et~al.(2019)Foerster, Song, Hughes, Burch, Dunning,
  Whiteson, Botvinick and Bowling}]{foerster2019bayesian}
\bibinfo{author}{Foerster, J.}, \bibinfo{author}{Song, F.},
  \bibinfo{author}{Hughes, E.}, \bibinfo{author}{Burch, N.},
  \bibinfo{author}{Dunning, I.}, \bibinfo{author}{Whiteson, S.},
  \bibinfo{author}{Botvinick, M.}, \bibinfo{author}{Bowling, M.},
  \bibinfo{year}{2019}.
\newblock \bibinfo{title}{Bayesian action decoder for deep multi-agent
  reinforcement learning}, in: \bibinfo{booktitle}{International Conference on
  Machine Learning}, \bibinfo{organization}{PMLR}. pp.
  \bibinfo{pages}{1942--1951}.
\bibitem[{Hansen et~al.(2004)Hansen, Bernstein and
  Zilberstein}]{POSGs2004hansen}
\bibinfo{author}{Hansen, E.A.}, \bibinfo{author}{Bernstein, D.S.},
  \bibinfo{author}{Zilberstein, S.}, \bibinfo{year}{2004}.
\newblock \bibinfo{title}{Dynamic programming for partially observable
  stochastic games}, in: \bibinfo{booktitle}{AAAI}, pp.
  \bibinfo{pages}{709--715}.
\bibitem[{Hart and Mas-Colell(2000)}]{hart2000simple}
\bibinfo{author}{Hart, S.}, \bibinfo{author}{Mas-Colell, A.},
  \bibinfo{year}{2000}.
\newblock \bibinfo{title}{A simple adaptive procedure leading to correlated
  equilibrium}.
\newblock \bibinfo{journal}{Econometrica} \bibinfo{volume}{68},
  \bibinfo{pages}{1127--1150}.
\bibitem[{Hor{\'a}k and Bo{\v{s}}ansk{\'y}(2019)}]{horak_public}
\bibinfo{author}{Hor{\'a}k, K.}, \bibinfo{author}{Bo{\v{s}}ansk{\'y}, B.},
  \bibinfo{year}{2019}.
\newblock \bibinfo{title}{Solving partially observable stochastic games with
  public observations}, in: \bibinfo{booktitle}{Proceedings of the AAAI
  Conference on Artificial Intelligence}, pp. \bibinfo{pages}{2029--2036}.
\bibitem[{Jakobsen et~al.(2016)Jakobsen, S{\o}rensen and
  Conitzer}]{timeability}
\bibinfo{author}{Jakobsen, S.K.}, \bibinfo{author}{S{\o}rensen, T.B.},
  \bibinfo{author}{Conitzer, V.}, \bibinfo{year}{2016}.
\newblock \bibinfo{title}{Timeability of extensive-form games}, in:
  \bibinfo{booktitle}{Proceedings of the 2016 ACM Conference on Innovations in
  Theoretical Computer Science}, \bibinfo{organization}{ACM}. pp.
  \bibinfo{pages}{191--199}.
\bibitem[{Johanson et~al.(2011)Johanson, Waugh, Bowling and
  Zinkevich}]{accelerated_BR}
\bibinfo{author}{Johanson, M.}, \bibinfo{author}{Waugh, K.},
  \bibinfo{author}{Bowling, M.}, \bibinfo{author}{Zinkevich, M.},
  \bibinfo{year}{2011}.
\newblock \bibinfo{title}{Accelerating best response calculation in large
  extensive games}, in: \bibinfo{booktitle}{IJCAI}, pp.
  \bibinfo{pages}{258--265}.
\bibitem[{Kova{\v{r}}{\'i}k and Lis{\'y}(2019)}]{ObsBasedEFGmodel}
\bibinfo{author}{Kova{\v{r}}{\'i}k, V.}, \bibinfo{author}{Lis{\'y}, V.},
  \bibinfo{year}{2019}.
\newblock \bibinfo{title}{Problems with the {EFG} formalism: A solution attempt
  using observations}.
\newblock \bibinfo{journal}{arXiv preprint arXiv:1906.06291} .
\bibitem[{Kroer and Sandholm(2016)}]{CRSWFG}
\bibinfo{author}{Kroer, C.}, \bibinfo{author}{Sandholm, T.},
  \bibinfo{year}{2016}.
\newblock \bibinfo{title}{Imperfect-recall abstractions with bounds in games},
  in: \bibinfo{booktitle}{Proceedings of the 2016 ACM Conference on Economics
  and Computation}, pp. \bibinfo{pages}{459--476}.
\bibitem[{Kuhn(1950)}]{Kuhn1950}
\bibinfo{author}{Kuhn, H.W.}, \bibinfo{year}{1950}.
\newblock \bibinfo{title}{A simplified two-person poker}.
\newblock \bibinfo{journal}{Contributions to the Theory of Games 1} .
\bibitem[{Lagoudakis and Parr(2002)}]{factoredVF2}
\bibinfo{author}{Lagoudakis, M.G.}, \bibinfo{author}{Parr, R.},
  \bibinfo{year}{2002}.
\newblock \bibinfo{title}{Learning in zero-sum team {M}arkov games using
  factored value functions}.
\newblock \bibinfo{journal}{Advances in Neural Information Processing Systems}
  \bibinfo{volume}{15}, \bibinfo{pages}{1659--1666}.
\bibitem[{Lanctot et~al.(2012)Lanctot, Burch, Zinkevich, Bowling and
  Gibson}]{lanctot2012}
\bibinfo{author}{Lanctot, M.}, \bibinfo{author}{Burch, N.},
  \bibinfo{author}{Zinkevich, M.}, \bibinfo{author}{Bowling, M.},
  \bibinfo{author}{Gibson, R.G.}, \bibinfo{year}{2012}.
\newblock \bibinfo{title}{No-regret learning in extensive-form games with
  imperfect recall}, in: \bibinfo{booktitle}{Proceedings of the 29th
  International Conference on Machine Learning (ICML-12)}, pp.
  \bibinfo{pages}{65--72}.
\bibitem[{Lerer et~al.(2020)Lerer, Hu, Foerster and Brown}]{lerer2020improving}
\bibinfo{author}{Lerer, A.}, \bibinfo{author}{Hu, H.},
  \bibinfo{author}{Foerster, J.N.}, \bibinfo{author}{Brown, N.},
  \bibinfo{year}{2020}.
\newblock \bibinfo{title}{Improving policies via search in cooperative
  partially observable games.}, in: \bibinfo{booktitle}{AAAI}, pp.
  \bibinfo{pages}{7187--7194}.
\bibitem[{Li et~al.(2020)Li, Hu, Zhang, Wang, Zhou, Qi and Song}]{li2020regret}
\bibinfo{author}{Li, H.}, \bibinfo{author}{Hu, K.}, \bibinfo{author}{Zhang,
  S.}, \bibinfo{author}{Wang, L.}, \bibinfo{author}{Zhou, J.},
  \bibinfo{author}{Qi, Y.}, \bibinfo{author}{Song, L.}, \bibinfo{year}{2020}.
\newblock \bibinfo{title}{Regret minimization via novel vectorized sampling
  policies and exploration}.
\newblock \bibinfo{journal}{preprint} \URLprefix
  \url{http://aaai-rlg.mlanctot.info/papers/AAAI20-RLG_paper_14.pdf}.
  \bibinfo{note}{accessed: August 5th, 2020}.
\bibitem[{Morav{\v{c}}{\'\i}k et~al.(2017)Morav{\v{c}}{\'\i}k, Schmid, Burch,
  Lis{\'y}, Morrill, Bard, Davis, Waugh, Johanson and Bowling}]{DeepStack}
\bibinfo{author}{Morav{\v{c}}{\'\i}k, M.}, \bibinfo{author}{Schmid, M.},
  \bibinfo{author}{Burch, N.}, \bibinfo{author}{Lis{\'y}, V.},
  \bibinfo{author}{Morrill, D.}, \bibinfo{author}{Bard, N.},
  \bibinfo{author}{Davis, T.}, \bibinfo{author}{Waugh, K.},
  \bibinfo{author}{Johanson, M.}, \bibinfo{author}{Bowling, M.},
  \bibinfo{year}{2017}.
\newblock \bibinfo{title}{Deepstack: Expert-level artificial intelligence in
  heads-up no-limit poker}.
\newblock \bibinfo{journal}{Science} \bibinfo{volume}{356},
  \bibinfo{pages}{508--513}.
\bibitem[{Moravcik et~al.(2016)Moravcik, Schmid, Ha, Hladik and
  Gaukrodger}]{moravcik2016refining}
\bibinfo{author}{Moravcik, M.}, \bibinfo{author}{Schmid, M.},
  \bibinfo{author}{Ha, K.}, \bibinfo{author}{Hladik, M.},
  \bibinfo{author}{Gaukrodger, S.J.}, \bibinfo{year}{2016}.
\newblock \bibinfo{title}{Refining subgames in large imperfect information
  games}, in: \bibinfo{booktitle}{Thirtieth AAAI Conference on Artificial
  Intelligence}, pp. \bibinfo{pages}{572--578}.
\bibitem[{Nayyar et~al.(2013)Nayyar, Mahajan and
  Teneketzis}]{nayyar2013decentralized}
\bibinfo{author}{Nayyar, A.}, \bibinfo{author}{Mahajan, A.},
  \bibinfo{author}{Teneketzis, D.}, \bibinfo{year}{2013}.
\newblock \bibinfo{title}{Decentralized stochastic control with partial history
  sharing: A common information approach}.
\newblock \bibinfo{journal}{IEEE Transactions on Automatic Control}
  \bibinfo{volume}{58}, \bibinfo{pages}{1644--1658}.
\bibitem[{von Neumann and Morgenstern(1953)}]{morgenstern1953theory}
\bibinfo{author}{von Neumann, J.}, \bibinfo{author}{Morgenstern, O.},
  \bibinfo{year}{1953}.
\newblock \bibinfo{title}{Theory of games and economic behavior}.
\newblock \bibinfo{publisher}{Princeton university press}.
\bibitem[{Nisan et~al.(2007)Nisan, Roughgarden, Tardos and
  Vazirani}]{nisan2007algorithmic}
\bibinfo{author}{Nisan, N.}, \bibinfo{author}{Roughgarden, T.},
  \bibinfo{author}{Tardos, E.}, \bibinfo{author}{Vazirani, V.V.},
  \bibinfo{year}{2007}.
\newblock \bibinfo{title}{Algorithmic game theory}.
\newblock \bibinfo{publisher}{Cambridge university press}.
\bibitem[{Oliehoek(2013)}]{oliehoek2013sufficient}
\bibinfo{author}{Oliehoek, F.A.}, \bibinfo{year}{2013}.
\newblock \bibinfo{title}{Sufficient plan-time statistics for decentralized
  {POMDP}s}, in: \bibinfo{booktitle}{Twenty-Third International Joint
  Conference on Artificial Intelligence}, pp. \bibinfo{pages}{302--308}.
\bibitem[{Oliehoek et~al.(2008)Oliehoek, Spaan, Vlassis and
  Whiteson}]{factoredDecPOMDP1}
\bibinfo{author}{Oliehoek, F.A.}, \bibinfo{author}{Spaan, M.T.},
  \bibinfo{author}{Vlassis, N.}, \bibinfo{author}{Whiteson, S.},
  \bibinfo{year}{2008}.
\newblock \bibinfo{title}{Exploiting locality of interaction in factored
  {Dec-POMDPs}}, in: \bibinfo{booktitle}{Int. Joint Conf. on Autonomous Agents
  and Multi-Agent Systems}, pp. \bibinfo{pages}{517--524}.
\bibitem[{Oliehoek et~al.(2009)Oliehoek, Whiteson and
  Spaan}]{oliehoek2009lossless}
\bibinfo{author}{Oliehoek, F.A.}, \bibinfo{author}{Whiteson, S.},
  \bibinfo{author}{Spaan, M.T.}, \bibinfo{year}{2009}.
\newblock \bibinfo{title}{Lossless clustering of histories in decentralized
  {POMDP}s}, in: \bibinfo{booktitle}{Proceedings of The 8th International
  Conference on Autonomous Agents and Multiagent Systems-Volume 1},
  \bibinfo{organization}{International Foundation for Autonomous Agents and
  Multiagent Systems}. pp. \bibinfo{pages}{577--584}.
\bibitem[{Oliehoek et~al.(2013)Oliehoek, Whiteson, Spaan
  et~al.}]{factoredDecPOMDP2}
\bibinfo{author}{Oliehoek, F.A.}, \bibinfo{author}{Whiteson, S.},
  \bibinfo{author}{Spaan, M.T.}, et~al., \bibinfo{year}{2013}.
\newblock \bibinfo{title}{Approximate solutions for factored {Dec-POMDP}s with
  many agents.}, in: \bibinfo{booktitle}{AAMAS}, pp. \bibinfo{pages}{563--570}.
\bibitem[{Peshkin et~al.(2000)Peshkin, Kim, Meuleau and
  Kaelbling}]{peshkin2000learning}
\bibinfo{author}{Peshkin, L.}, \bibinfo{author}{Kim, K.E.},
  \bibinfo{author}{Meuleau, N.}, \bibinfo{author}{Kaelbling, L.P.},
  \bibinfo{year}{2000}.
\newblock \bibinfo{title}{Learning to cooperate via policy search}, in:
  \bibinfo{booktitle}{Proceedings of the Sixteenth conference on Uncertainty in
  artificial intelligence}, \bibinfo{organization}{Morgan Kaufmann Publishers
  Inc.}. pp. \bibinfo{pages}{489--496}.
\bibitem[{Russell(2019)}]{humanCompatible}
\bibinfo{author}{Russell, S.}, \bibinfo{year}{2019}.
\newblock \bibinfo{title}{Human compatible: Artificial intelligence and the
  problem of control}.
\newblock \bibinfo{publisher}{Penguin}.
\bibitem[{Schrittwieser et~al.(2019)Schrittwieser, Antonoglou, Hubert,
  Simonyan, Sifre, Schmitt, Guez, Lockhart, Hassabis, Graepel et~al.}]{muZero}
\bibinfo{author}{Schrittwieser, J.}, \bibinfo{author}{Antonoglou, I.},
  \bibinfo{author}{Hubert, T.}, \bibinfo{author}{Simonyan, K.},
  \bibinfo{author}{Sifre, L.}, \bibinfo{author}{Schmitt, S.},
  \bibinfo{author}{Guez, A.}, \bibinfo{author}{Lockhart, E.},
  \bibinfo{author}{Hassabis, D.}, \bibinfo{author}{Graepel, T.}, et~al.,
  \bibinfo{year}{2019}.
\newblock \bibinfo{title}{Mastering atari, {G}o, chess and shogi by planning
  with a learned model}.
\newblock \bibinfo{journal}{arXiv preprint arXiv:1911.08265} .
\bibitem[{Seitz et~al.(2019)Seitz, Kova{\v{r}}{\'\i}k, Lis{\'y}, Rudolf, Sun
  and Ha}]{seitzDLS}
\bibinfo{author}{Seitz, D.}, \bibinfo{author}{Kova{\v{r}}{\'\i}k, V.},
  \bibinfo{author}{Lis{\'y}, V.}, \bibinfo{author}{Rudolf, J.},
  \bibinfo{author}{Sun, S.}, \bibinfo{author}{Ha, K.}, \bibinfo{year}{2019}.
\newblock \bibinfo{title}{Value functions for depth-limited solving in
  imperfect-information games beyond poker}.
\newblock \bibinfo{journal}{arXiv preprint arXiv:1906.06412} .
\bibitem[{Shoham and Leyton-Brown(2008)}]{MAS}
\bibinfo{author}{Shoham, Y.}, \bibinfo{author}{Leyton-Brown, K.},
  \bibinfo{year}{2008}.
\newblock \bibinfo{title}{Multiagent systems: Algorithmic, game-theoretic, and
  logical foundations}.
\newblock \bibinfo{publisher}{Cambridge University Press}.
\bibitem[{Silver et~al.(2017)Silver, Schrittwieser, Simonyan, Antonoglou,
  Huang, Guez, Hubert, Baker, Lai, Bolton et~al.}]{alphaZero}
\bibinfo{author}{Silver, D.}, \bibinfo{author}{Schrittwieser, J.},
  \bibinfo{author}{Simonyan, K.}, \bibinfo{author}{Antonoglou, I.},
  \bibinfo{author}{Huang, A.}, \bibinfo{author}{Guez, A.},
  \bibinfo{author}{Hubert, T.}, \bibinfo{author}{Baker, L.},
  \bibinfo{author}{Lai, M.}, \bibinfo{author}{Bolton, A.}, et~al.,
  \bibinfo{year}{2017}.
\newblock \bibinfo{title}{Mastering the game of {G}o without human knowledge}.
\newblock \bibinfo{journal}{Nature} \bibinfo{volume}{550},
  \bibinfo{pages}{354}.
\bibitem[{Spaan et~al.(2008)Spaan, Oliehoek and Vlassis}]{spaan2008multiagent}
\bibinfo{author}{Spaan, M.T.}, \bibinfo{author}{Oliehoek, F.A.},
  \bibinfo{author}{Vlassis, N.}, \bibinfo{year}{2008}.
\newblock \bibinfo{title}{Multiagent planning under uncertainty with stochastic
  communication delays}, in: \bibinfo{booktitle}{338 Proceedings of the
  Eighteenth International Conference on Automated Planning and Scheduling
  (ICAPS 2008)}, pp. \bibinfo{pages}{338--345}.
\bibitem[{Srinivasan et~al.(2018)Srinivasan, Lanctot, Zambaldi, P{\'e}rolat,
  Tuyls, Munos and Bowling}]{srinivasan2018actor}
\bibinfo{author}{Srinivasan, S.}, \bibinfo{author}{Lanctot, M.},
  \bibinfo{author}{Zambaldi, V.}, \bibinfo{author}{P{\'e}rolat, J.},
  \bibinfo{author}{Tuyls, K.}, \bibinfo{author}{Munos, R.},
  \bibinfo{author}{Bowling, M.}, \bibinfo{year}{2018}.
\newblock \bibinfo{title}{Actor-critic policy optimization in partially
  observable multiagent environments}.
\newblock \bibinfo{journal}{arXiv preprint arXiv:1810.09026} .
\bibitem[{{\v{S}}ustr et~al.(2019){\v{S}}ustr, Kova{\v{r}}{\'\i}k and
  Lis{\'y}}]{MCCR}
\bibinfo{author}{{\v{S}}ustr, M.}, \bibinfo{author}{Kova{\v{r}}{\'\i}k, V.},
  \bibinfo{author}{Lis{\'y}, V.}, \bibinfo{year}{2019}.
\newblock \bibinfo{title}{{M}onte {C}arlo continual resolving for online
  strategy computation in imperfect information games}, in:
  \bibinfo{booktitle}{Proceedings of the 18th International Conference on
  Autonomous Agents and MultiAgent Systems},
  \bibinfo{organization}{International Foundation for Autonomous Agents and
  Multiagent Systems}. pp. \bibinfo{pages}{224--232}.
\bibitem[{Wichardt(2008)}]{wichardt2008}
\bibinfo{author}{Wichardt, P.C.}, \bibinfo{year}{2008}.
\newblock \bibinfo{title}{{Existence of Nash Equilibria in Finite Extensive
  Form Games with Imperfect Recall: A Counterexample}}.
\newblock \bibinfo{journal}{Games and Economic Behavior} \bibinfo{volume}{63},
  \bibinfo{pages}{366--369}.
\bibitem[{Wikipedia()}]{comm_know_wiki}
Wikipedia, .
\newblock \bibinfo{title}{Common knowledge (logic)}.
\newblock
  \bibinfo{howpublished}{\url{https://en.wikipedia.org/wiki/Common_knowledge_(logic)}}.
\newblock \bibinfo{note}{Accessed: August 5th, 2020}.
\bibitem[{de~Witt et~al.(2019)de~Witt, Foerster, Farquhar, Torr, Boehmer and
  Whiteson}]{de2019multi}
\bibinfo{author}{de~Witt, C.S.}, \bibinfo{author}{Foerster, J.},
  \bibinfo{author}{Farquhar, G.}, \bibinfo{author}{Torr, P.},
  \bibinfo{author}{Boehmer, W.}, \bibinfo{author}{Whiteson, S.},
  \bibinfo{year}{2019}.
\newblock \bibinfo{title}{Multi-agent common knowledge reinforcement learning},
  in: \bibinfo{booktitle}{Advances in Neural Information Processing Systems},
  pp. \bibinfo{pages}{9927--9939}.
\bibitem[{Zinkevich et~al.(2007)Zinkevich, Johanson, Bowling and
  Piccione}]{07nips-cfr}
\bibinfo{author}{Zinkevich, M.}, \bibinfo{author}{Johanson, M.},
  \bibinfo{author}{Bowling, M.}, \bibinfo{author}{Piccione, C.},
  \bibinfo{year}{2007}.
\newblock \bibinfo{title}{Regret minimization in games with incomplete
  information}, in: \bibinfo{booktitle}{Advances in Neural Information
  Processing Systems 20 (NIPS)}, pp. \bibinfo{pages}{905--912}.
\bibitem[{Zinkevich et~al.(2008)Zinkevich, Johanson, Bowling and
  Piccione}]{CFR}
\bibinfo{author}{Zinkevich, M.}, \bibinfo{author}{Johanson, M.},
  \bibinfo{author}{Bowling, M.}, \bibinfo{author}{Piccione, C.},
  \bibinfo{year}{2008}.
\newblock \bibinfo{title}{Regret minimization in games with incomplete
  information}, in: \bibinfo{booktitle}{Advances in neural information
  processing systems}, pp. \bibinfo{pages}{1729--1736}.

\end{thebibliography}

\appendix
    \section*{Appendix}
    \renewcommand{\appendixname}{{}}

\section{Sub-game Decomposition}\label{sec:app:decomposition}

In this section, we first discuss several different ways in which one can look at \modelAbbrev{}s and prove their compatibility (Section~\ref{sec:app:view-compatibility}).
Afterwards, we show how to formalize the notion of a subgame in a way that makes subgames a special case of the general \modelAbbrev{} definition.

\subsection{The Compatibility of Different Views of \modelAbbrev{}s}\label{sec:app:view-compatibility}

Recall that $\mc H$ denotes the set of histories, $\mc S_i$ the set of information states, and $\pubTree$ the set of public states.
When the \modelAbbrev{} model is known, a history $h$ conveys more information than the corresponding information state $s_i \in \mc S_i$, which in turns conveys more information than the corresponding public state. In this sense, the mappings $s_i : \mc H \to \mc S_i$, $\pub : \mc S_i \to \pubTree$ and $\pub : \mc H \to \pubTree$ can be viewed as projections, and we have ``$\pub^{\mc S_i} \circ s_i = \pub^{\mc H}$'' (see Figure~\ref{fig:kuhn_fog} for illustration).
\modelAbbrev{}s can also be studied using their extensive-form representation, where information and public states are replaced by infosets and public sets.
In this manner, each information- (resp. public-) \textit{state} can be viewed as a ``name'' or ``identifier'' for the corresponding information (public) \textit{set}.

Since the mappings $I_i : \mc S_i \to \mc I_i$ and $\pubSet : \pubTree \to \mc \pubSet$ are bijections, we can use $I_i \mapsto s_i(I_i)$ and $\pubSet \mapsto \pub (\pubSet)$ to denote the corresponding inverse mappings.
Since it is possible to compose many of these mappings together, we overload this notation using the ``target\_object\_type(starting\_object)'' convention --- for example, we set $I_i(h) := I_i(s_i(h))$ and $\pubSet (I_i) := (\pubSet \circ \pub \circ s_i)(I_i) = $ the (uniquely defined) public set which contains the given information set $I_i$.
Similarly, we can identify each public state $s \in \pubTree$ with the corresponding collection $\mc S_i(s) := \{ s_i \in \mc S_i \mid \pub (s_i) = s \}$ of information states or information sets $\mc I_i(s) := \{ I \in \mc I_i \mid \pub (I) = s \}$.

The following proposition then implies that we can safely view (and search trough) \modelAbbrev{}s on the level of histories, information states, or public states, and all these views are equivalent.
(For a graphical illustration, see Figure~\ref{fig:kuhn_fog}, which shows the \modelAbbrev{} representation of Kuhn poker \cite{Kuhn1950}.)

\begin{restatable}[Equivalent definitions of \modelAbbrev{} subgame trees]{proposition}{subgames}\label{prop:subgame}
For a history $h_0 \in \mc H$, $s=\pub(h_0)$ and $H\subset \mc H$, the following ``definitions of a subgame tree'' are equivalent:
\begin{enumerate}[label=(\roman*)]
    \item $H$ is the smallest set containing $h_0$ and closed under descendants in $\mc H$ and membership within public sets.
        \label{case:closure}
    \item $H = \{ g \in \mc H \mid g \textnormal{ extends some } h \in \pubSet(s) \}$.
        \label{case:histories}
    \item $H = \bigcup \{ I_i(s_i') \mid s_i' \textnormal{ extends some } s_i \in \mc S_i(s) \}$.
    \item $H = G(s) = \bigcup \{ \pubSet(s') \mid s' \textnormal{ extends } s \}$.
\end{enumerate}
Moreover, for such $H$ we have:
\begin{itemize}
    \item[$(v)$] $H$ is closed under descendants in $\mc H$ and membership within infosets.\footnote{However, $\pubSet(h)$ isn't necessarily \defword{the smallest} set containing $h$ closed under the membership within infosets.
(That would require working with all information that is common knowledge, not just the public part. However, it is the smallest such set that a model-free agent can use.)
As a result, $H$ needs not be the smallest set satisfying $(iv)$.}
\end{itemize}
\end{restatable}


\begin{proof}
The ``moreover'' part follows from, e.g., $(i)$, because closure under the membership within public states automatically implies closure under the membership within information sets. Indeed, if two histories belong to the same information set, they produce the same sequence of private and public observations (and actions). Consequently, they trivially also produce the same sequence of public observations, i.e., they belong to the same public state.

Denote by $H_1$, $H_2$, $H_3$, and $G(s)$ the sets from $(i)$, $(ii)$, $(iii)$ and $(iv)$.
The inclusion $H_1 \supset H_2$ is trivial.
Note also that the second identity in $(iv)$ always holds since $g \in \pubSet(s')$ $\iff$ $\pub(g)=s'$.

For $H_2 \supset G(s)$, suppose that we have $\pub(h_0) = (O^0, \dots, O^t)$ and $g' = (s^0, a^0, \dots, s^{t'})  \in G(s)$.
Since by definition of $G(s)$, $\pub(g')$ extends $\pub(h_0)$, we have $\pub(g) = (O^1, \dots, O^t, \dots, O^{t'})$ for some $t'\geq t$.
The history $g := (s^0, a^0, \dots, s^t) \sqsubset g'$ witnesses that $g' \in H_2$.

To see that $G(s) \supset H_1$, note that $G(s)$ is \emph{a} set containing $h_0$ and closed under both the membership within public states (by definition) and descendants in $\mc H$ (since $h' \sqsupset h_0$ implies that $\pub(h')$ is an extension of $\pub(h_0)$).
This yields $G(s) \supset H_1$.

The identity between $H_3$ and $G(s)$ holds since for any $s'\in \pubTree$, we have
\begin{equation*}
\pubSet(s') = \bigcup \{ I_i(s_i') \mid s_i' \in \mc S_i(s') \} .
\end{equation*}
\end{proof}



\begin{figure}[htb]
\centering
\begin{tikzpicture}[scale=1]
\sneakingGamePartition
\end{tikzpicture}
\caption{An \textit{EFG} where \ref{case:closure} and \ref{case:histories} from Proposition~\ref{prop:subgame} are not equivalent. (Taken from our earlier technical report \cite{ObsBasedEFGmodel}.)
}
\label{fig:sneak}
\end{figure}

In particular, this proposition implies that we can safely view (and search trough) subgames on the level of histories, information states, or public states, and all these views are equivalent.

Critically, this result does not hold in EFGs since some of them might fail the implication $(ii)\implies (i)$ --- indeed, consider the EFG from Figure~\ref{fig:sneak} with ``public states = histories at the same level''.
Moreover, even ``well-behaved'' EFGs have the drawback of not having a notion of ``agent's information'' outside of the agent's turn.
This implies that in EFGs, the conditions $(iii)$ and $(iv)$ are not defined at all.
While there have been attempts to remedy this \cite{CFR-D}, these approaches did not work as one would intuitively expect or hope (see also Section~\ref{sec:EFG-problems}).

\subsection{Formal Construction of Subgames}

Suppose that $G$ is a \modelAbbrev{} and $\beta = (\pub, r)$ is a public belief state in $G$.
We will show how to formally construct the subgame $G(\pub, r)$ in a way that it is both consistent with our intuitions and formally satisfies the definition of a \modelAbbrev{}.

By $S := \pubSet(\pub)$, we denote the public set corresponding to $\pub$.
Suppose that $r$ is of the form $r = (P^\pi_i(h))_{h \in S, \, i=1,\dots,N,c}$, i.e., it contains the information about reach probabilities of each player\edit{,
which} allows us to compute the normalized joint reach probabilities $P^\pi(h|S) = P^\pi(h)/\sum_{g\in S} P^\pi(g)$ at $\pub$.\footnote{This is possible even when $P^\pi(S)=0$ \cite{seitzDLS}.}
We will construct $G(\beta)$ by taking the \modelAbbrev{} $G$, giving it a new initial world-state $\initState_\beta$, and defining how transitioning from $\initState_\beta$ affects rewards and observations.

This new initial state works as follows.
No players are active at $\initState_\beta$.
To emulate the situation described by $\beta$, a history $h\in S$ is internally sampled from $P^\pi(\cdot|S)$.
This causes the game to transition to the world-state $w(h)$ (where $w(h)$ denotes the last world-state encountered along the history $h$).
Upon this transition, each player receives the reward $\mc R_i(h)$ that they would receive along this history and privately observes the corresponding infostate $s_i(h)$.\footnote{Formally, this means that rewards and observations at $\initState_\beta$ are stochastic. This comes at no loss in generality, since the same result could be achieved by adding auxiliary world-states $w_h$, $h\in S$, and keeping rewards and observations deterministic.}
Finally, the public observation thus generated consists not only of $\pub$ (which we would expect), but additionally contains the range $r$.

\bigskip

The reason for including the range in the public observation is the following.
The purpose of subgames is to allow us to run various algorithms only on the subgame while still getting identical results as if the computation was run on the whole game.
Since many algorithms, including CFR, work with the reach probabilities of $P^\pi_i(h)$ of individual players, the subgame should provide accurate information about these numbers.
However, since $G(\beta)$ emulates the players' actions via a \textit{chance} node, every reach probability that was originally due to some \textit{player} is now formally transformed into a chance reach probability.
As a result, naively running an algorithm on $G(\beta)$ would produce different results, likely undesirably so.
On the other hand, if an algorithm is provided the range $r$ through the public observation, it can replace these wrong numbers $P^\pi_i(h)$ with the ones given by $r$.
Whenever applying an algorithm to a subgame, we will assume that this is the case.

Admittedly, the approach does seem somewhat unprincipled.
We thus welcome suggestions for improvements in this direction.
In the meantime, the cleanest way we are aware of, for making the approach consistent with the stand definition, is the following:

\begin{remark}[Defining counterfactual values in terms of an initial range]
We now describe how to make the definition of counterfactual values \textit{in subgames} consistent with their standard version.
First, observe that any \textit{standard} \modelAbbrev{} can be modelled as a \modelAbbrev{} \textit{subgame} that corresponds to a public belief state $\beta = (\pub, r)$, where $\pub$ is an empty observation and $r$ is a range that assigns probability 1 to the initial world-state $\initState$.
In subgames, we then re-define reach probabilities as follows:
if $g$ is a history in the root public set and $h$ is some extension of $g$, then $P^\pi_i(h)$ is the product of $P^\pi_i(g)$ given by the initial range and $P^\pi_i(g,h)$ given by the strategy in the subgame.
Counterfactual values can then be defined in terms of this upgraded definition.
The resulting notion will have the desired properties in actual subgames while retaining their original behaviour in standard games.
\end{remark}
\section{Proofs}
     \AugEFGize*

\begin{proof}
The perfect-recall of $\EF{G}$ holds because each player's information states satisfy an analogous property by definition. The ``no thick public states'' property holds because in the default \modelAbbrev{} model, each player receives a non-trivial public observation on each transition.
For the moreover part, we turn $\mc H$ into the set of world-states, each $\mc A(h)$ into the set of legal actions, chance strategy $\pi_c$ into the transition function, $u$ into the rewards (for terminal states only), and infosets/public sets into private/public observations.
The extensive form of this game will coincide with the original augmented EFG.
\end{proof}

\EFGize*

\begin{proof}
Since we already have Theorem~\ref{thm:FOSG-to-aug-EFG}, it suffices to show that the mapping $\texttt{ForgetNonActingI}(\cdot)$ sends augmented perfect-recall EFGs with no thick public sets to 1-timeable perfect-recall classical EFGs.
This holds because we can define an exact deterministic timing for $E'$ as $\tau(h) := $ the number of public sets encountered on the way to $h$. (If the timing wasn't exact, $E$ couldn't have perfect recall.)

For the moreover part, we need to start with a classical EFG and extend its $\mc I'_i$ into partitions defined on the full $\mc H$, and define the corresponding public partition.
For $i\in \mc N$, we define a function $\varphi$ on $\mc H$ as:
\begin{itemize}
    \item $\varphi(h) := I$ if $h \in I \in \mc I'_i$,
    \item $\varphi(ha) := Ia$ if $h\in I \in \mc I'_i$, $a\in \mc A(h)$, and $p(ha) \neq i$,
    \item $\varphi(h) := \emptyset$ if $h$ is the root of $\mc H$ (but $p(h)\neq i$),
    \item $\varphi(h) := \varphi(g) k $ if $\varphi(h)$ hasn't been defined in one of the above steps and $k$ is the distance to the nearest ancestor $g$ of $h$ for which $\varphi(g)$ is defined.
\end{itemize}
We then define $\mc I_i$ as the equivalence class of histories that have the same value of $\varphi$.
A public partition $\pubSetTree$ compatible with $\mc I_1,\dots,\mc I_N$ can be constructed by starting with $\bigcup_{i\in \mc N} \mc I_i$ and whenever two sets intersect each other, replacing them by their union.
It is trivially true that restricting $\mc I_i$ to $\{ h \mid p(h) = i \}$ yields $\mc I'_i$.
Similarly, the definition of $\varphi$ ensures that $\mc I_i$ contains no thick infosets (which implies that neither does $\pubSetTree$, given how it was defined).
The only non-trivial part is thus verifying that each $\mc I_i$ has perfect recall.
However, this follows from the assumption that the original EFG was 1-timeable. Indeed, the only ``extra'' information that $\mc I_i$ contains over $\mc I'_i$ is the number of histories between every two subsequent elements of $\mc I'_i$. By the 1-timeability assumption, this value must be the same among all members of any given infoset $I\in \mc I'_i$.
\end{proof}

\begin{example}[Serialization of \modelAbbrev{}s]\label{ex:serialization}
For a \modelAbbrev{} $G$, the (serial) \modelAbbrev{} $\texttt{Serialize}(G) := \left< \mc N, \mc W', p', \initState, \mc A, \mc T', \mc R', \mc O' \right>$ can be obtained by the following construction:
\begin{itemize}
 \item For every $w\in \mc W$, $\mc W'$ formally contains states $w_{i,a_{<i}}$ and $w_{c,a}$, where $i\in p(w)$, $a_{<i} \in \prod_{j\in p(w), j<i} \mc A_j(w)$, and $a\in \prod_{j\in p(w)} \mc A_j(w)$.
 \item The player function is defined as $p'(w_{i,a<i}) := \{i\}$ and $p'(w) := \emptyset$.
 \item The legal actions are defined as in $G$, i.e., we have $\mc A_i(w_{i,a_{<i}}) = \mc A_i(w)$.
 \item Taking $a_i$ at $w_{i,<i}$ always transitions the game to $w_{i+1,a_{<i}a_i}$ (resp. to $w_{j,a_{<i}a_i}$ for the first $i<j\in \mc N$ that acts at $w$, or to $w_{c,a_{<i}a_i}$ if there is no such $j$).
 The \textit{noop} action always taken at $w_{c,a}$ translates to $w'$ (resp. to the corresponding $w'_{\min p(w'),\emptyset}$) in accordance to $\mc T(w,a)$.
 \item Upon transitioning from $w_{c,a}$ to some $w'_{(\cdot)}$, the game generates rewards $\mc R'(w_{c,a},noop) := \mc R(w,a)$ and observations $\mc O'(w_{c,a},noop,w'_{(\cdot)}) := \mc O(w,a)$. Other transitions only produce trivial observations.
\end{itemize}
\end{example}
\section{The proof of the \modelAbbrev{} version of CFR} \label{app:CFR}

In this section, we give the proof of our earlier claim that the CFR algorithm in \modelAbbrev{}s behaves identically to its original EFG version.

\CFR*

\begin{proof}
Let $G$ and $E'$ be a \modelAbbrev{} as in the proposition.
Firstly, note that the EFG-ization of $G$ induces a mapping between information states of $G$ (resp., those where $i$ acts) and information sets of $E'$.
This translates to a perfect correspondence between strategies in $G$ and strategies in $E'$.
Moreover, the counterfactual reach probability of $s_i \in \mc S_i$ will always be equal to the counterfactual reach probability of the corresponding infoset $I \in \mc I'_i$.
By the assumption made on $G$, we can define $\mc R_i : \mc I'_i \to \R$ in such a way that $\mc R_i(h) = \mc R_i(I)$ whenever $h$ is a history in $G$ for which $s_i(h)$ corresponds to $I$.
The correspondence is less straightforward for histories, since each history $h\in \mc H^G$ in $G$ might correspond to multiple histories $\Phi(h) = \{ ha_{<i} \mid a_{<i} \in \prod_{j < i} \mc A_i^G(h) \} \subset \mc H^{E'}$.
However, for any $h\in \mc H^G$, we will always have $\mc R_i(h) + v^\pi_i(h) = \sum_{g\in \Phi(h)} P^\pi(h|\Phi(h)) \textnormal{EFG-}v_i^\pi(g) $
(where $\textnormal{EFG}-v_i^\pi(g)$ denotes the expected utility conditional on using $\pi$ and currently being in $g$).
(Intuitively, $i$ doesn't know the actions of their opponents, but from their point of view, it does not matter whether these actions have already been taken or whether the opponents are still deciding.)
It follows that for any strategy $\pi$, the information states and infosets that correspond to each other will always satisfy
$v^\pi_i(s_i) = \textnormal{EFG-}v_i^\pi(I) - \mc R_i(I)$ and $q^\pi_i(s_i,a_i) = \textnormal{EFG-}q_i^\pi(I,a_i) - \mc R_i(I)$.
Since $P^\pi_{-i}(s_i) = P^\pi_{-i}(I)$, the same will be true for counterfactual values.
This implies that the counterfactual regrets are also identical, which concludes the proof.
\end{proof}

\section{Bounding the Size Increase of Making EFGs 1-timeable}\label{sec:app:padding_size}

In this section, we show that padding a timeable game to become 1-timeable will increase its size at most quadratically and that this bound is tight.

\begin{lemma}\label{lem:int_val_timing}
Any timeable EFG has a an exact deterministic timing with values in $\{ 0, \dots, | \mc H | -1 \}$.
\end{lemma}

\begin{proof}
Let $\psi : \mc H \to [0,\infty)$ be an exact deterministic timing.
We enumerate $\{ t_0, \dots, t_N \} := \{ \psi(h) \mid h \in \mc H \}$ s.t. $t_n < t_{n+1}$.
We necessarily have $N \leq | \mc H | -1$.
For $h\in \mc H$, we denote $\varphi(h) := n$ for the integer $n$ satisfying $\psi(h) = t_n$.
By definition of $\varphi$, we have $\varphi(g) = \varphi(h)$ iff $\psi(g) = \psi(h)$.
Since $\psi$ is a deterministic timing, we have $h \sqsubseteq h'$ $\implies$ $\varphi(h) \leq \varphi(h')$.
It follows that $\varphi$ is an exact deterministic timing that satisfies $\{ \varphi(h) \mid h \in \mc H \} = \{ 0, \dots, N \} \subset \{ 0, \dots, | \mc H | -1 \}$.
\end{proof}

For a fixed integer-valued timing $\varphi$ in an EFG $E$ and a history $ha \in \mc H$ (where $h\in \mc H$, $a \in \mc A^E(h)$), we denote by $\tau_{\varphi}(ha) := (\varphi(ha) - \varphi(h) ) - 1$ the number of ``missing'' time-stamps between $ha$ and its parent.
In this sense, the ``length'' of the transition from $h$ to $ha$ via $a$ is $ 1 + \tau_{\varphi}(ha)$.

\begin{lemma}[Padding an EFG with non-unit transitions]
\label{lem:padding_formula}
For any EFG $E$ with exact deterministic integer-valued timing $\varphi$, there exists a strategically equivalent 1-timeable EFG $E'$ with $| \mc H_{E'} | = | \mc H | + \sum_{ha} \tau_{\varphi}(ha)$.
\end{lemma}

\begin{proof}
Let $E$ and $\varphi$ be as in the lemma.
For every $h\in \mc H_E$ and $a \in \mc A^E(h)$, we formally add to $\mc H_{E'}$ the nodes $h$, $ha_n$ for $n=1,\dots,\tau_{\varphi}(ha)$, and $ha$.
(If $h$ as already been added as $h'a'$ for its parent $h'$, we don't add it for the second time.)
In $E'$-histories of the form $h$ and $ha$, both the acting player and available actions are as in $E$.
The chance player acts in all the nodes of the form $ha_n$, and he only has a single \emph{noop} action.
The transitions in $E'$ work as follows: $h \overset{a}{\curvearrowright} ha_1 \overset{noop}{\curvearrowright} \dots \overset{noop}{\curvearrowright} ha_n \overset{noop}{\curvearrowright} ha$.
The remaining objects in $E'$ (player set, information partitions, utilities) are as in $E$.

The identity mappings $h \in \mc H_E \mapsto h \in \mc H_{E'}$ and $a \in \mc A^E(h) \mapsto a \in \mc A^{E'}(h)$ induce an isomorphism between strategies in $E$ and $E'$, which shows that $E$ and $E'$ are strategically equivalent.
Moreover, the deterministic timing $\psi : h \in \mc H_{E'} \mapsto \textnormal{ the number of actions taken between the root of $\mc H_{E'}$ and $h$}$ is exact in $E'$.
We conclude the proof by noting that the length of every transition in $E'$ is 1, and that $\mc H_{E'}$ has the desired size.
\end{proof}

Together with Lemma~\ref{lem:int_val_timing}, this gives the following corollary:
\begin{corollary}[The padding is quadratic]\label{cor:padding}
Any timeable EFG $E$ admits a strategically equivalent 1-timeable EFG $E'$ with $| \mc H_{E'} | \leq | \mc H |^2$.
\end{corollary}

Figure~\ref{fig:quadratic_increase} depicts a domain where the increase is quadratic.

\begin{figure}
\centering
\scalebox{0.75}{
    \quadraticPaddingGameA
}
\scalebox{0.75}{
    \quadraticPaddingGameB
}
\caption{A game where Lemma~\ref{lem:padding_formula} causes a quadratic growth.
Left: An EFG $E$ where transitioning between $h_n$ and $g_n$ takes $N-n$ units of time.
Right: The padded version of $E$, where transitions take precisely 1 unit of time.
Upwards/downwards facing triangles denote which player acts --- maximizer (pl.\,1) or minimizer (pl.\,2).
}
\label{fig:quadratic_increase}\label{fig:padding}
\end{figure}

\end{document}